\documentclass[sigconf]{acmart}

\usepackage{multirow}
\usepackage{setspace}
\usepackage{lipsum}
\usepackage{enumitem}
\usepackage{booktabs,tabularx}
\newcolumntype{Y}{>{\centering\arraybackslash}X}

\AtBeginDocument{%
  }

\newcommand\blfootnote[1]{
\begingroup
\renewcommand\thefootnote{}\footnote{#1}
\addtocounter{footnote}{-1}
\endgroup
}
\copyrightyear{2023}
\acmYear{2023}
\setcopyright{rightsretained}
\acmConference[MM '23]{Proceedings of the 31st ACM International Conference on Multimedia}{October 29-November 3, 2023}{Ottawa, ON, Canada}
\acmBooktitle{Proceedings of the 31st ACM International Conference on Multimedia (MM '23), October 29-November 3, 2023, Ottawa, ON, Canada}
\acmDOI{10.1145/3581783.3611801}
\acmISBN{979-8-4007-0108-5/23/10}

\makeatletter
\gdef\@copyrightpermission{
  \begin{minipage}{0.3\columnwidth}
   \href{https://creativecommons.org/licenses/by/4.0/}{\includegraphics[width=0.90\textwidth]{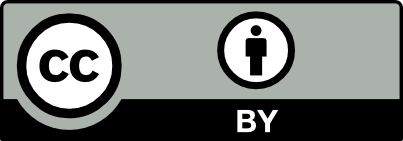}}
  \end{minipage}\hfill
  \begin{minipage}{0.7\columnwidth}
   \href{https://creativecommons.org/licenses/by/4.0/}{This work is licensed under a Creative Commons Attribution International 4.0 License.}
  \end{minipage}
  \vspace{5pt}
}
\makeatother

\acmSubmissionID{465}

\begin{document}
% \fancyhead{}

\title{Towards Robust Real-Time Scene Text Detection: From Semantic to Instance Representation Learning}

% \author{Xugong Qin$^{1*}$, Pengyuan Lyu$^{3*}$, Chengquan Zhang$^{3}$, Yu Zhou$^{2,4\dagger}$,\\ Kun Yao$^{3}$, Peng Zhang$^{1}$, Hailun Lin$^{2}$, and Weiping Wang$^{2}$
% }
% \affiliation{
%   \institution{
%   $^1$School of Cyber Science and Engineering, Nanjing University of Science and Technology, Nanjing, China\\
%   $^2$Institute of Information Engineering, Chinese Academy of Sciences, Beijing, China\\
%   $^3$Department of Computer Vision Technology (VIS), Baidu Inc., Beijing, China\\
%   $^4$School of Cyber Security, University of Chinese Academy of Sciences, Beijing, China
%   }
%   \streetaddress{}
%   \city{}
%   \state{}
%   \country{}
%   \postcode{}
% }
% \email{{qinxugong, zhang_peng}@njust.edu.cn, {zhouyu, linhailun, wangweiping}@iie.ac.cn}
% \email{{lvpengyuan, zhangchengquan, yaokun01}@baidu.com}

\author{Xugong Qin$^{*}$}
\affiliation{
  \institution{School of Cyber Science and Engineering, Nanjing University of Science and Technology}
  \country{}
}
\email{qinxugong@njust.edu.cn}

\author{Pengyuan Lyu$^{*}$}
\affiliation{
  \institution{Department of Computer Vision Technology (VIS), Baidu Inc.}
  \country{}
}
\email{lvpengyuan@baidu.com}

\author{Chengquan Zhang}
\affiliation{
  \institution{Department of Computer Vision Technology (VIS), Baidu Inc.}
  \country{}
}
\email{zhangchengquan@baidu.com}

\author{Yu Zhou$^{\dagger}$}
\affiliation{
  \institution{Institute of Information Engineering, Chinese Academy of Sciences}
  \institution{School of Cyber Security, University of Chinese Academy of Sciences}
  \country{}
}
\email{zhouyu@iie.ac.cn}

\author{Kun Yao}
\affiliation{
  \institution{Department of Computer Vision Technology (VIS), Baidu Inc.}
  \country{}
}
\email{yaokun01@baidu.com}

\author{Peng Zhang}
\affiliation{
  \institution{School of Cyber Science and Engineering, Nanjing University of Science and Technology}
  \country{}
}
\email{zhang_peng@njust.edu.cn}

\author{Hailun Lin}
\affiliation{
  \institution{Institute of Information Engineering, Chinese Academy of Sciences}
  \country{}
}
\email{linhailun@iie.ac.cn}

\author{Weiping Wang}
\affiliation{
  \institution{Institute of Information Engineering, Chinese Academy of Sciences}
  \country{}
}
\email{wangweiping@iie.ac.cn}

\renewcommand{\shortauthors}{Xugong Qin et al.}


\begin{abstract}
Due to the flexible representation of arbitrary-shaped scene text and simple pipeline, bottom-up segmentation-based methods begin to be mainstream in real-time scene text detection. Despite great progress, these methods show deficiencies in robustness and still suffer from false positives and instance adhesion. Different from existing methods which integrate multiple-granularity features or multiple outputs, we resort to the perspective of representation learning in which auxiliary tasks are utilized to enable the encoder to jointly learn robust features with the main task of per-pixel classification during optimization. For semantic representation learning, we propose global-dense semantic contrast (GDSC), in which a vector is extracted for global semantic representation, then used to perform element-wise contrast with the dense grid features. To learn instance-aware representation, we propose to combine top-down modeling (TDM) with the bottom-up framework to provide implicit instance-level clues for the encoder. With the proposed GDSC and TDM, the encoder network learns stronger representation without introducing any parameters and computations during inference. Equipped with a very light decoder, the detector can achieve more robust real-time scene text detection. Experimental results on four public datasets show that the proposed method can outperform or be comparable to the state-of-the-art on both accuracy and speed. Specifically, the proposed method achieves 87.2\% F-measure with 48.2 FPS on Total-Text and 89.6\% F-measure with 36.9 FPS on MSRA-TD500 on a single GeForce RTX 2080 Ti GPU. 
\end{abstract}





\maketitle

\begin{small}
\begin{spacing}
1
\textbf{ACM Reference Format:}

\noindent Xugong Qin, Pengyuan Lyu, Chengquan Zhang, Yu Zhou, Kun Yao, Peng Zhang, Hailun Lin, and Weiping Wang. 2023. Towards Robust Real-Time Scene Text Detection: From Semantic to Instance Representation Learning. In \textit{Proceedings of the 31st ACM International Conference on Multimedia (MM ’23), October 29–November 3, 2023, Ottawa, ON, Canada.} ACM, New York, NY, USA, 10 pages. https://doi.org/10.1145/3581783.3611801
\end{spacing}
\end{small}

\blfootnote{*Equal contribution. The work was mainly done at IIE, CAS when Xugong Qin was an intern at Baidu Inc.. $\dagger$Yu Zhou is the corresponding author.}

%-------------------------------------------------------------------------
\vspace{-15px}
\begin{figure}[!htb]
\begin{center}
\includegraphics[width=0.7\linewidth]{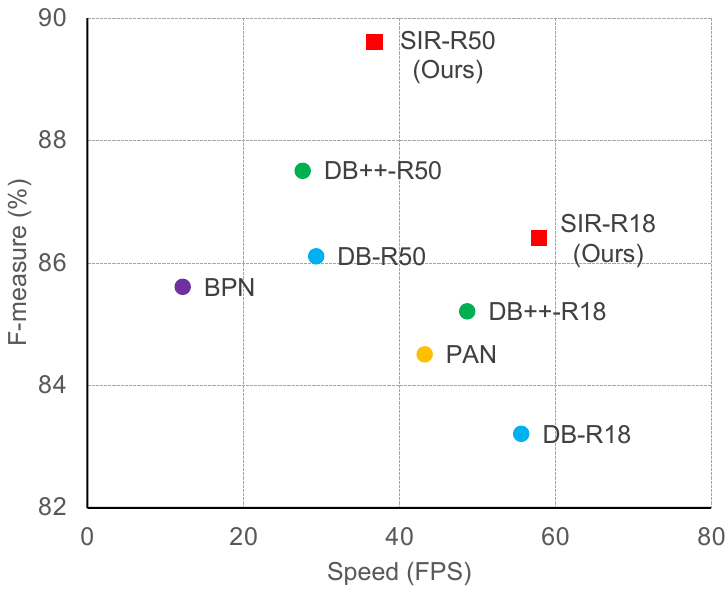}
\end{center}
\vspace{-10px}
\caption{The comparison of recent SOTA methods on the MSRA-TD500 dataset. SIR achieves the best accuracy/speed trade-off. All results listed here are reproduced from the corresponding official implementation on the same device. ``R18'' and ``R50'' denote ResNet-18 and ResNet-50.}
\label{fig:speed_accuracy}
\end{figure}

\section{Introduction}
\label{sec:intro}

Optical character recognition has attracted much attention due to its practical applications, e.g., scene understanding \cite{wang2022kmfgr,zgy2021mm,zgy2022pr}, blind navigation, and document analysis \cite{long2022hiertext,shen2023divide}. As an essential part of the text reading system \cite{wang2022tpsnet,wj2022textblock}, scene text detection (STD) is widely studied. Significant progress has been witnessed in the past decades \cite{ye2015survey,zhu2016fcs_survey,long2021era,yao2014strokelets,huang2014robust,tian2016CTPN,gupta2016synthetic,yao2016holistic,shi2017SegLink,lyu2018corner,xing2019CCN,liao2020db,wang2022TSTD,wan2021STKD,song2022VLPT,xue2022oCLIP,yu2023TCM,guo2021icann,guo2022icme}. However, STD remains challenging due to significant variations in scale, orientation, and aspect ratio as well as curved shape.

Due to the flexible representation of arbitrary-shaped scene text, segmentation-based methods have become mainstream and popular in academic and industrial communities. Top-down methods \cite{he2017maskrcnn,wang2020solov2} perform foreground/background segmentation based on the ``locate-then-seg'' pipeline. To make it simple and efficient, the whole image can be parsed as text/non-text region \cite{long2015FCN,yao2016holistic,Zhang2016FCNTD} from the bottom-up view. However, distinguishing different instances becomes an essential problem since scene text could be born crowded. Among methods settling this, shrinking text regions \cite{zhou2017east,wu2017self,wang2019PSENet,wang2019PAN,liao2020db} become the most prevalent way. Nevertheless, these methods are with different implementations, which makes it hard to compare them in one framework. Moreover, the performance is generally inferior to top-down methods, which are not robust to instance adhesion and false positives issues. 

\vspace{-10px}
\begin{figure}[!htb]
\begin{center}
    \includegraphics[width=0.9\linewidth]{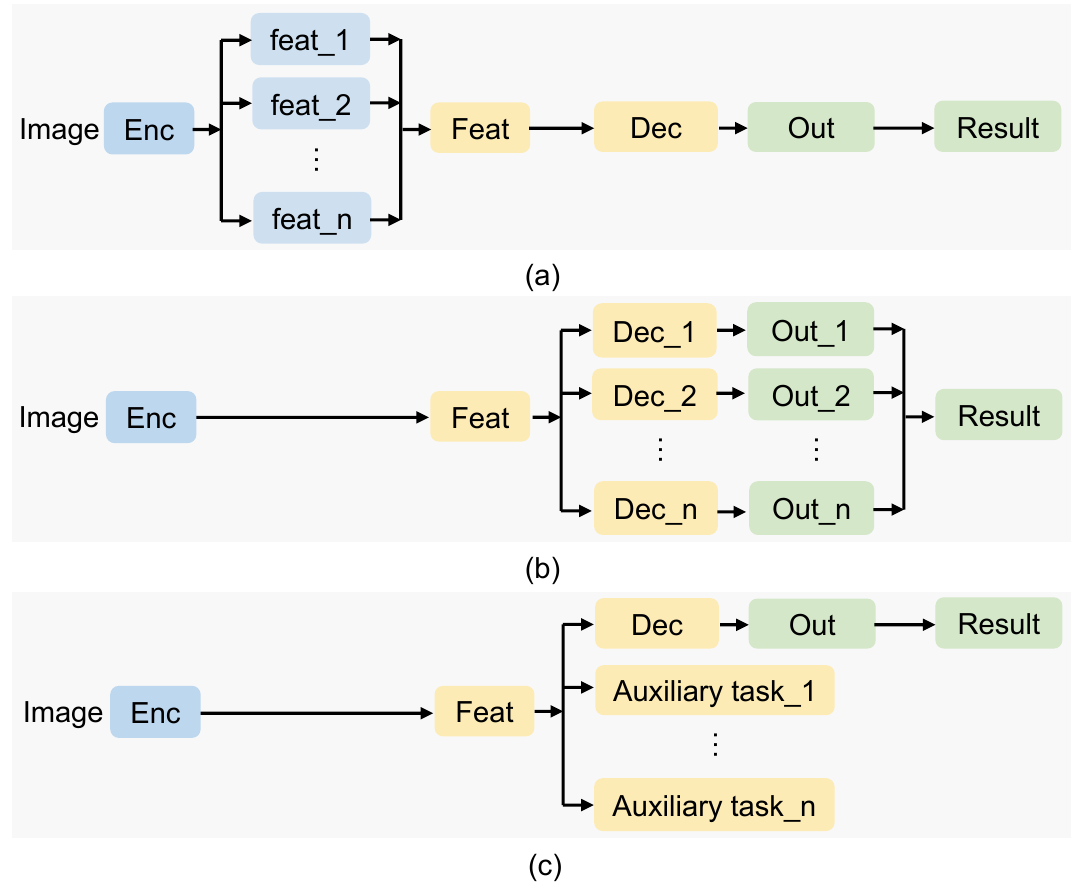}
\end{center}
\vspace{-10px}
\caption{Different learning paradigms for robust scene text detection within an encoder-decoder framework. (a): Fusing multi-granularity features \cite{ye2020textfusenet} for rich representation. (b): Fusing multiple outputs from different modeling perspectives \cite{xie2019SPCNet,liu2019BDN}. (c): Representation learning with auxiliary tasks (ours). ``Enc'', ``Feat'', ``Dec'', and ``Out'' denote encoder, feature, decoder and output respectively.}
\label{fig:learn_paradigm}
\end{figure}
\vspace{-10px}

The learning paradigms of existing robust STD methods can be roughly summarized into two categories as shown in Fig. \ref{fig:learn_paradigm} (a) and (b). As shown in Fig. \ref{fig:learn_paradigm} (a), one kind of method is fusing multi-granularity features to obtain rich representation. TextfuseNet \cite{ye2020textfusenet} is a typical work in which features from semantic segmentation, word level, and character level are extracted and fused. The other category method follows the pipeline listed in Fig. \ref{fig:learn_paradigm} (b), which fuses multiple outputs from different modeling perspectives to re-correct the confidence of predictions \cite{xie2019SPCNet,liu2019BDN} to suppress the false positives. The above methods all focus on top-down methods, and enjoy improved performance but are slowed by the burdened pipelines. To achieve robust real-time scene text detection, we choose popular bottom-up segmentation which is efficient and accurate yet still not fully exploited. To avoid bringing additional computation and parameters in testing, we resort to the perspective of representation learning and develop a different learning paradigm in which auxiliary tasks are paralleled with the decoder in training to help the encoder learn robust representation. 

However, two issues matter in robust representation learning. One is the insufficiency of semantic-level modeling. Specifically, long-range relation modeling is only achieved by a stack of locally connected convolution layers, which makes it inefficient to relate distant pixels and maintain semantic consistency across the whole image. The other lies in the absence of instance-level modeling. Since scene text detection is essentially an instance-level task. Though the shrinking text region is widely adopted to distinguish different text instances, it works in a bottom-up way. We argue that the lack of global instance-level concepts which naturally exist in top-down methods makes it one of the key reasons for the inferior performance. 

To address the issues, we first compose a unified baseline model which unifies existing methods into a weighted three-channel segmentation problem. For semantic representation learning, we propose global-dense semantic contrast (GDSC) as an auxiliary task in which a global semantic vector is extracted and contrasted to the dense feature vectors, which not only helps to maintain semantic consistency but also is efficient in computation. Different from SAE \cite{tian2019SAE} and PAN \cite{wang2019PAN} focus on pulling features from the same instances, and pushing away features from different instances, GDSC operates on the semantic level which keeps the intra-class (text) distance small and inter-class distances (text/non-text) large.
For instance-level representation learning, we propose to integrate top-down modeling (TDM) with bottom-up segmentation into one framework, inspired by GTC \cite{hu2020gtc} which uses a top-down attention mechanism \cite{shi2018aster} to supervise bottom-up CTC \cite{shi2016crnn} methods in text recognition \cite{shi2016crnn,shi2018aster,du2022svtr,qiao2020gcan,qiao2020seed,qiao2021pimnet}. Specifically, the widely adapted two-stage detection framework Faster R-CNN is utilized to perform regional recognition tasks, which provide instance-level information during the encoder learning. 

In this work, we propose a robust real-time scene text detection framework, namely \textbf{SIR}, which explores from \textbf{S}emantic to \textbf{I}nstance \textbf{R}epresentation learning within a bottom-up segmentation framework.
The main contributions of the paper are as follows:
\begin{itemize}
\item We propose a new framework for robust real-time scene text detection which utilizes auxiliary tasks to help the encoder learn robust representation without introducing any computations or parameters during inference.
\item We revisit existing works, unify existing segmentation-based methods into a simple weighted three-channel cross-entropy loss and propose a strong baseline model.
\item Global-dense semantic contrast is proposed for effective semantic representation learning. We exploit top-down modeling in a segmentation-based framework and find it significantly boosts the performance of the bottom-up segmentation pipeline.
\item Experiments on four public datasets including horizontal, multi-oriented, and curved text demonstrate the effectiveness and robustness of the proposed method on real-time scene text detection.
\end{itemize}

%-------------------------------------------------------------------------
\section{Related Work}
According to different modeling perspectives, existing scene text detection methods can be divided into bottom-up (BU) and top-down (TD) methods. 
Considering different modeling granularity, methods can be divided into pixel-based and region-based/contour-based methods, which correspond to segmentation-based (SEG) and regression-based (REG) methods in previous work. To better summarize existing works, we provide a detailed categorization of existing STD methods: TD-REG, TD-SEG, BU-REG, and BU-SEG. 

\textbf{TD-REG} methods \cite{zhou2017east,liu2017DMPNet,he2017deepregression,he2017SSTD,liao2018textboxes++,wang2018ITN,liao2018RRD,xue2018border,wang2019SAST,Liu2020ABCNet,qin2021fc2rn,tang2022FSG,wang2022tpsnet} regress the geometry of text bounding boxes, e.g., the horizontal rectangles \cite{ren2015fasterrcnn}, oriented rectangles \cite{ma2018RRPN,zhou2017east}, quadrilaterals \cite{zhou2017east,liu2017DMPNet,liao2018textboxes++,wang2018ITN,qin2021fc2rn}, polygons \cite{liu2019curved} or parametric curves \cite{Liu2020ABCNet,Wang2020textray,zhu2021fce,wang2022tpsnet}, and contours \cite{dai2021pcr,zhang2021textbpn}. 
These methods are modified from the modern general detectors \cite{liu2016ssd,ren2015fasterrcnn} which perform relative position regression based on anchor boxes or anchor points with one-stage prediction or multi-stage refinement.  

\textbf{TD-SEG} methods \cite{he2017maskrcnn,yang2018inceptext,liao2019masktextspotterv2,liu2019BDN,qin2019wsctd,ye2020textfusenet,Wang2020contournet,liu2019maskTTD,qin2021mayor,he2021most} perform instance segmentation based on region features or grid features. Mask R-CNN \cite{he2017maskrcnn} build a mask head upon Faster R-CNN \cite{ren2015fasterrcnn}, which performs per-pixel classification in a given region of interest (RoI) whether it belongs to the instance corresponding to the RoI. To alleviate errors accumulated from multiple stages, one-stage instance segmentation \cite{wang2020solov2} is proposed to directly produce an instance-level mask. However, TD methods require multiple stages or deep stack convolution due to the limitation of the receptive field. On the contrary, BU methods alleviate the demand by predicting local units, which gives a more flexible representation. 

\textbf{BU-REG} methods \cite{tian2016CTPN,shi2017SegLink,long2018textsnake,tang2019seglink++,xue2019msr,zhang2020DRRG,zhou2020crnet,ma2021relatext} learns to regress local geometry which is also based on predefined anchors. Local bounding boxes \cite{tian2016CTPN,shi2017SegLink,long2018textsnake,tang2019seglink++,zhang2020DRRG,ma2021relatext} or boundary points \cite{xue2019msr,zhou2020crnet,sheng2021centripetaltext} are predicted in BU-SEG methods, which can handle long text well. However, these kinds of methods usually require complex post-processing and tend to suffer from false positives.  

\textbf{BU-SEG} methods \cite{Zhang2016FCNTD,he2017cias,wu2017self,deng2018pixellink,xu2019textfield,tian2019SAE,wang2019PSENet,wang2019PAN,liao2020db,sheng2021centripetaltext,liao2022db++} regard text detection as semantic segmentation problem and perform per-pixel classification on the input image. However, different from the general semantic segmentation tasks, different text instances must be separable on the output semantic maps to obtain the bounding box of each text instance. Thus shrinking text regions is widely adopted in BU-SEG methods.
To reconstruct the text regions from the text kernel, dilating text contour \cite{wu2017self,liao2020db}, progressive scale expansion \cite{wang2019PSENet}, and embedding clustering \cite{wang2019PAN,tian2019SAE} are proposed.
Compared with TD methods, BU methods need to link the local unit predictions to generate instance-level results. Due to the lack of instance-level supervision, BU-SEG methods not only struggled to deal with false positives but also suffered from instance adhesion problems.

Generally speaking, methods can be divided into multiple categories, e.g., Mask R-CNN \cite{he2017maskrcnn} involves TD-Reg and TD-SEG stages, and Corner \cite{lyu2018corner} includes BU-REG corner localization and BU-SEG position-aware segmentation. The proposed method is also inspired by the complementary of different modeling perspectives.

Real-time scene text detection \cite{zhou2017east,wang2019PAN,liao2020db,liao2022db++} not only cares about the performance but also the inference speed. 
Existing work mainly focuses on light structures with elaborate designs. Efficient feature fusion is widely studied in 
EAST \cite{zhou2017east}, PAN \cite{wang2019PAN}, DB \cite{liao2020db}, DB++ \cite{liao2022db++} as well as compact model design and efficient learning mechanisms. Due to shallow network design and simple supervision, existing real-time methods suffer from insufficient representation and inaccurate detection results.
In this work, different from these methods, we aim to improve the representation learned by the encoder without introducing parameters and computation during inference, which enables robust real-time scene text detection. 

%------------------------------------------------------------------------
\section{Methodology}

The proposed method consists of three parts: a BU-SEG baseline, GDSC, and TDM. We first briefly describe the overall pipeline of the proposed method and then detail these parts. The label generation and optimization objectives are presented at last.

\begin{figure*}[!htb]
\begin{center}
\includegraphics[width=0.6\linewidth]{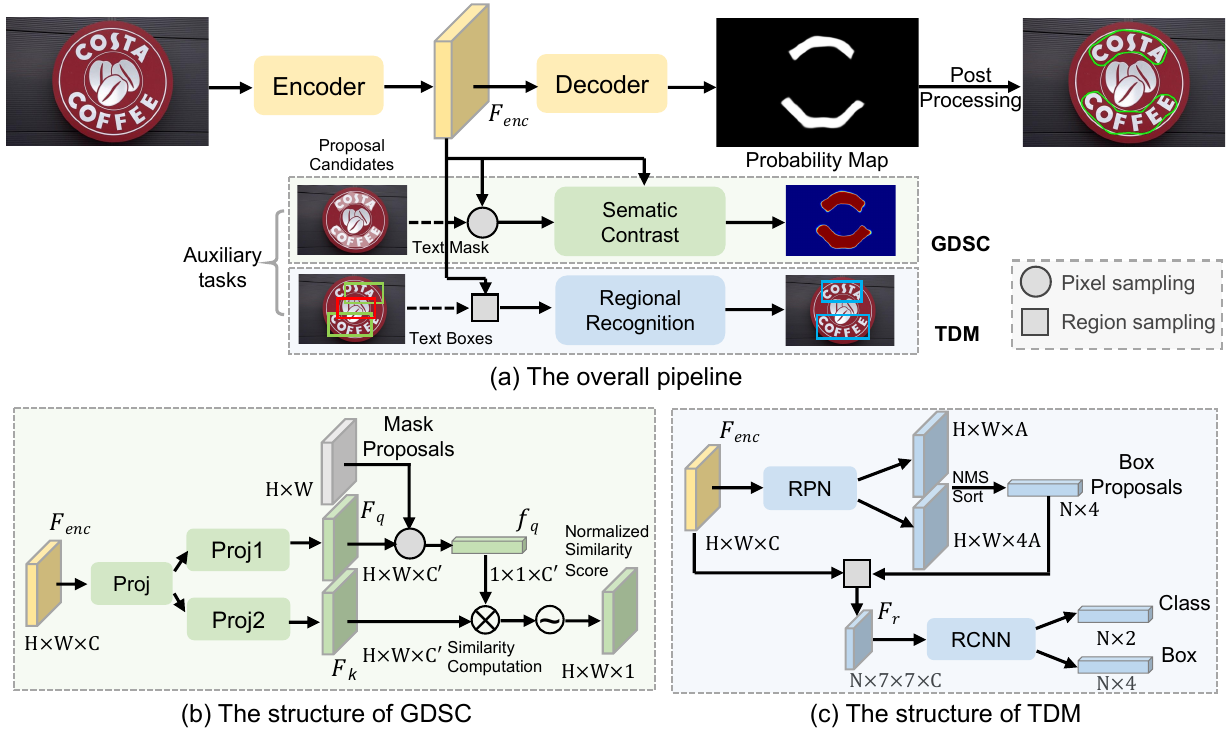}
\end{center}
\vspace{-10px}
\caption{The overall pipeline of SIR. It mainly contains one BU-SEG main network and two auxiliary tasks including GDSC and TDM. The auxiliary tasks only work in training.}
\label{fig:struct}
\end{figure*}

The architecture of SIR is illustrated in Fig. \ref{fig:struct}, which consists of the main and auxiliary tasks. The main task is a BU-SEG task which is a typical encoder-decoder-based framework. The auxiliary tasks include GDSC and TDM, which are in parallel with the decoder to promote the encoder in training and removed during inference. Given an input image, the feature $F_{enc} \in {\mathbb R}^{H \times W \times C}$ is obtained through the encoder network and then decoded to produce the probability map via a lightweight decoder. In training, besides the main segmentation task, GDSC and TDM perform as the auxiliary tasks which take the encoded feature $F_{enc}$ as the input and propagate the gradients back to the encoder. The GDSC and TDM promote representation learning from the semantic and instance levels, providing strong complements to the man task without bringing burdens during inference.

%-------------------------------------------------------------------------
\subsection{A Unified Bottom-Up Segmentation Baseline}

In this subsection, we explore building a unified segmentation baseline for different implementations of shrinking text regions and fairly compare different variants \cite{yao2016holistic,xue2018border,liao2020db} within the same framework. 
As shown in Fig. \ref{fig:struct} (a), the baseline model follows an encoder-decoder segmentation framework.

\noindent\textbf{Encoder.} The encoder consists of the backbone and the neck. ResNet \cite{he2016resnet} with deformable convolution \cite{zhu2019DCNv2} is used as the backbone, followed by an FPN \cite{lin2017FPN,liao2020db,liao2022db++} and the adaptive scale fusion module \cite{liao2022db++}. The structure of the encoder is the same as that in \cite{liao2022db++}. The output dimension of the encoder is 256 and the spatial dimension is $\frac{1}{4}$ of the input image.

\noindent\textbf{Decoder.} The decoder part includes a 3$\times$3 convolution with 64 channels, followed by a BN and a ReLU, and a convolution layer for prediction. To be compatible with the border class \cite{wu2017self}, we set the channel of prediction convolution to 3, which represents non-text, text kernel, and text border. The output is up-sampled to the same scale as the input image and a ``softmax'' function is used to normalize the probabilities of different classes.   

\noindent\textbf{Loss Function.} Due to a serious imbalance between the foreground and background classes, class-aware balancing is widely used. We use a simple class-aware weighting strategy that is inversely proportional to the numbers of the corresponding classes. The segmentation loss can be defined as a weighted cross-entropy loss:
\begin{equation}
\mathcal{L}_{seg} = \frac{1}{HW}\sum_{i=1}^{H}\sum_{j=1}^{W}w_{ij}{\rm CE}(p_{ij}, y_{ij}),
\end{equation}
where $p$, $y$, and $w$ denote the prediction, the ground truth, and the corresponding weight. And the weight $w$ is defined as follows:
\begin{equation}
w_{ij} = {\mathbb W}^{\top}y_{ij},
\end{equation}
in which per pixel is weighted based on a class-aware weight $\mathbb W$ and can be defined as follows if taking border class into account:
\begin{equation}
{\mathbb W} = [1, \frac{N_{nt}}{N_{tk}}, \frac{N_{nt}}{N_{tb}}],
\end{equation}
or 
\begin{equation}
{\mathbb W} = [1, \frac{N_{nt}+N_{tb}}{N_{tk}}],
\end{equation}
if consider the border class as background,
where $N_{nt}$, $N_{tk}$, and $N_{tb}$ mean the number of pixels which belong to the non-text, text kernel, and text border regions.

\noindent\textbf{Supervision.} Different supervision within one model is achieved as follows. (1): Kernel/ non-kernel segmentation \cite{wang2019PAN,liao2020db,sheng2021centripetaltext,liao2022db++}. The probability of non-kernel is obtained by the sum of probabilities of the border and non-text. (2): (1) + class-aware weights (consider kernel/non-kernel only, denoted as W\_kn). (3): (1) + class-aware weights (kernel/border/non-text, W\_kbn). (4): (3) + explicit border and non-text supervision \cite{wu2017self}. (5): Text region segmentation \cite{yao2016holistic,ch2017total} with W\_kbn. The probability of text region is given out by summing the prediction from the kernel and border channels.

\noindent\textbf{Post-process.}
During inference, the probability map is first obtained by selecting the kernel class prediction from the output of the decoder and then binarized with a given threshold. Then connected components labeling (CCL) is performed on the binarized map. Finally, we obtain predicted polygons by dilating the contour of connected components with the Vatti clipping algorithm \cite{vatti1992vatti} following \cite{liao2020db,liao2022db++}.

Experiments with different supervision are performed on the ICDAR2015 (IC15) and Total-Text (TT) datasets for comparison. As shown in Table \ref{tab:baseline}, with W\_kn, kernel segmentation can be promoted from 74.1\% to 80.2\% F-measure on TT. However, we find a drop of 2.6\% on ICDAR2015 in which incident text occupies. Moreover, we observed, the kernel/non-kernel boundary is hard to decide with W\_kn. We find this phenomenon originates from two aspects. One is the imbalanced weight. The border region is taken as non-kernel which has a much smaller weight compared to the kernel region. When the border regions are explicitly weighted, the results are significantly improved (52.9\%->80.3\% on IC15, 80.2\%->82.4\% on TT). 
The other is the fussy semantic labels in which the border and non-text classes are coupled with W\_kn. The boundary pixels are hard to be correctly discriminated due to the fussy semantics of the hand-craft boundary class. When the border class is independently weighed and decoupled from the non-kernel class, the network is given clear semantic boundary information, leading to better results (80.3\%->81.2\% on IC15, 82.4\%->83.7\% on TT). The last row in Table \ref{tab:baseline} shows that the weighted three-channel classification is also compatible with the text region segmentation settings \cite{yao2016holistic,ch2017total}. 

\vspace{-5px}
\begin{table}[!htb]
\centering
\caption{Detection results with different settings of the BU-SEG baseline on ICDAR2015 and Total-Text. ``P'', ``R'', and ``F'' denote precision, recall, and F-measure respectively.}
\vspace{-10px}
% \small
\footnotesize
\begin{tabularx}{1.0\linewidth}{@{}llYYYYYY@{}}
\toprule 
\multirow{2}{*}{\#} & \multirow{2}{*}{Method} & \multicolumn{3}{c}{ICDAR2015}                 & \multicolumn{3}{c}{Total-Text}                \\ \cline{3-8} 
                       &                         & P             & R             & F             & P             & R             & F             \\ \hline
1                      & kernel                  & 66.3          & 47.7          & 55.5          & 85.9          & 65.2          & 74.1          \\
2                      & kernel + W\_kn             & 53.0          & 52.7          & 52.9          & 83.8          & 76.9          & 80.2          \\
3                      & kernel + W\_kbn            & 82.3          & \textbf{78.5} & 80.3          & 82.7          & \textbf{82.2} & 82.4          \\
4                      & kernel + W\_kbn + border     & \textbf{85.0} & 77.8          & \textbf{81.2} & \textbf{88.8} & 79.2          & \textbf{83.7} \\ 
5                      & text + W\_kbn              & 42.8          & 29.2          & 34.7          & 70.2          & 66.3          & 68.2          \\ \bottomrule
\end{tabularx}
\label{tab:baseline}
\end{table}
\vspace{-5px}

\subsection{Global-Dense Semantic Contrast}
Global-dense semantic contrast is proposed to enhance the semantic representation in a contrastive way. As shown in Fig. \ref{fig:struct} (b), given encoded feature $F_{enc} \in {\mathbb R}^{H \times W \times C}$, a shared project layer is first used to change the channel dimension to $C^{'}$, then two individual project heads are further utilized to embed the feature to obtain $F_{q} \in {\mathbb R}^{H \times W \times C^{'}}$ and $F_{k} \in {\mathbb R}^{H \times W \times C^{'}}$. The vector $f_q$ is extracted by feature sampling on $F_k$ according to the proposal mask $M \in {\mathbb R}^{H \times W}$, which aggregates semantic-rich features and averages to obtain the global semantic representation:
\begin{equation}
f_{q} = \frac{\sum_{x,y}F_k(x,y)\mathbb{I}[M(x,y)=1]}{\sum_{x,y}\mathbb{I}[M(x,y)=1]},
\end{equation}
in which $M$ is obtained by interpolation of the text region mask $T$ to the shape of $H \times W$.
We use a $3 \times 3$ convolution followed by a BN and a ReLU and $1 \times 1$ convolutions to implement the share project layer and the separated project heads.
$C^{'}$ is set to 128 in implementation.

After that, we measure the similarity between $f_q$ and each element $F(x,y)$ of $F_k$ by computing the inner product, which can be efficiently implemented with matrix multiplication. Then a normalized similarity score $S$ is further obtained by scaling the similarity to (0, 1) with the ``sigmoid'' function, which avoids introducing the hyper-parameter margin \cite{tian2019SAE,wang2019PAN} in metric learning:
\begin{equation}
S(x,y) = {\rm sig}(f_{q}^{\top}F_{k}(x,y)).
\end{equation} 
$S$ is up-sampled to the same scale as the input image.
We use the text region binary map $T$ to supervise the learning process with the dice loss \cite{milletari2016Vnet}, the loss can be defined as follows:
\begin{equation}
\mathcal{L}_{gdsc} = 1 - \frac{2\sum{T(x,y)S(x,y)}}{\sum{T(x,y)}+ \sum{S(x,y)}}.
\end{equation}
Since $T$ is a binary map, in which 1 represents text (+), 0 represents non-text (-). The above equation can be written as:
\begin{equation}
% M = sigmoid(F_{k}f_{q}^T))
\mathcal{L}_{gdsc} = 1 - \frac{2\sum_{k_+}{\rm sig}(f_{q}^{\top}k_+)}
{\sum{T} + \sum_{k_+}{\rm sig}(f_{q}^{\top}k_{+}) + \sum_{k_-}{\rm sig}(f_{q}^{\top}k_-)},
\end{equation}
where $\{k_+\}$ and $\{k_+\}$ satisfy:
\begin{equation}
\{k_+\} = \{F_k(x,y)|T(x,y)=1\},
\end{equation}
\begin{equation}
\{k_-\} = \{F_k(x,y)|T(x,y)=0\},
\end{equation}
which is a similar form to the typical InfoNCE \cite{oord2018cpc}, in which contrastive learning is performed between the global semantic representation and features within the same image.

GDSC performs dense contrast between a global semantic representation and the dense grid features, which can be viewed as a dynamic convolution \cite{jia2016DFN} in which the convolution parameters are conditioned on a global representation, enabling better semantic representation learning in an elegant way. 

\subsection{Combining Top-Down Modeling}
In the BU-SEG framework, text instances are obtained from CCL on the probability map during inference, which lacks the guidance of instance-level concepts in representation learning and is sensitive to false positives and text adhesion. To settle this issue, we introduce TDM through which the instance concept is naturally given out to the BU-SEG framework.

Without loss of generality, we choose the classical region-based TD modeling method Faster R-CNN \cite{ren2015fasterrcnn}. As shown in Fig. \ref{fig:struct} (c), given the encoded feature $F_{enc}$, TDM performs anchor-based dense supervision and proposal-based spare supervision at pixel-level and region-level respectively, enabling the encoder to learn instance-aware representation.

\noindent\textbf{Anchor-Based Dense Supervision.} From the view of BU-SEG, the encoded $F_{enc}$ is used to decode to produce a probability of each class. Considering from the TD perspective, each vector from the spatial position of $F_{enc} \in {\mathbb R}^{H \times W \times C}$ encoded the information of one area on the original input image. Associated with one predefined anchor per location as that in RPN \cite{ren2015fasterrcnn} and SSD \cite{liu2016ssd}, $F_{enc}$ can be utilized to produce the prediction of text instances by text/non-text classification and relative geometry regression, which provide instance-level information in a pixel-wise manner.

Standard RPN \cite{ren2015fasterrcnn} is used to provide anchor-based dense supervision. To process scene text of different scales, we perform consecutive max-pooling on $F_{enc}$ and construct a feature pyramid with scales of $\{\frac{1}{4}, \frac{1}{8}, \frac{1}{16}, \frac{1}{32}, \frac{1}{64}\}$ of the input image.

As stated in ATSS \cite{Zhang2020ATSS}, despite the way of encoding relative position information, the result of using the anchor box and anchor point are very close. If we don't take the interpolation into consideration, the supervision of the predicted probability map \cite{liao2020db,wang2019PAN} equals the supervision of the anchor point in EAST \cite{zhou2017east}, from which we can infer that anchor-based dense supervision has been performed in the BU-SEG framework on a certain degree.
When anchor-based dense supervision is applied, the instance concept is further strengthened.  

\noindent\textbf{Proposal-Based Sparse Supervision.}
Given the encoded feature $F_{enc}$ and a set of proposals, we first extract regional features $F_{r} \in {\mathbb R}^{N \times 7 \times 7 \times C}$ with RoI-Align \cite{he2017maskrcnn}, in which $N$ denote the numbers of proposals. Then regional recognition tasks including region-wise classification and regression are performed. Specifically, regional classification of text/non-text complements pixel-level classification with high-level clues. Besides classifying pixels into one semantic category, regional classification learns to distinguish whether the region contains a complete instance that is visually and semantically self-consistent. Regional regression learns to predict region-wise relative position, enabling the encoder to learn to identify the  boundary of text instances, which not only helps to promote the accuracy of semantic segmentation but also benefits obtaining complete instance contour by CCL.  

We use the fast R-CNN branch \cite{ren2015fasterrcnn} to implement the proposal-based sparse supervision. The proposals are obtained by sorting and performing NMS on the output of RPN, which proves to work quite well in practice.

\label{fig:vis}
\begin{figure*}[!htb]
\begin{center}
\includegraphics[width=0.7\linewidth]{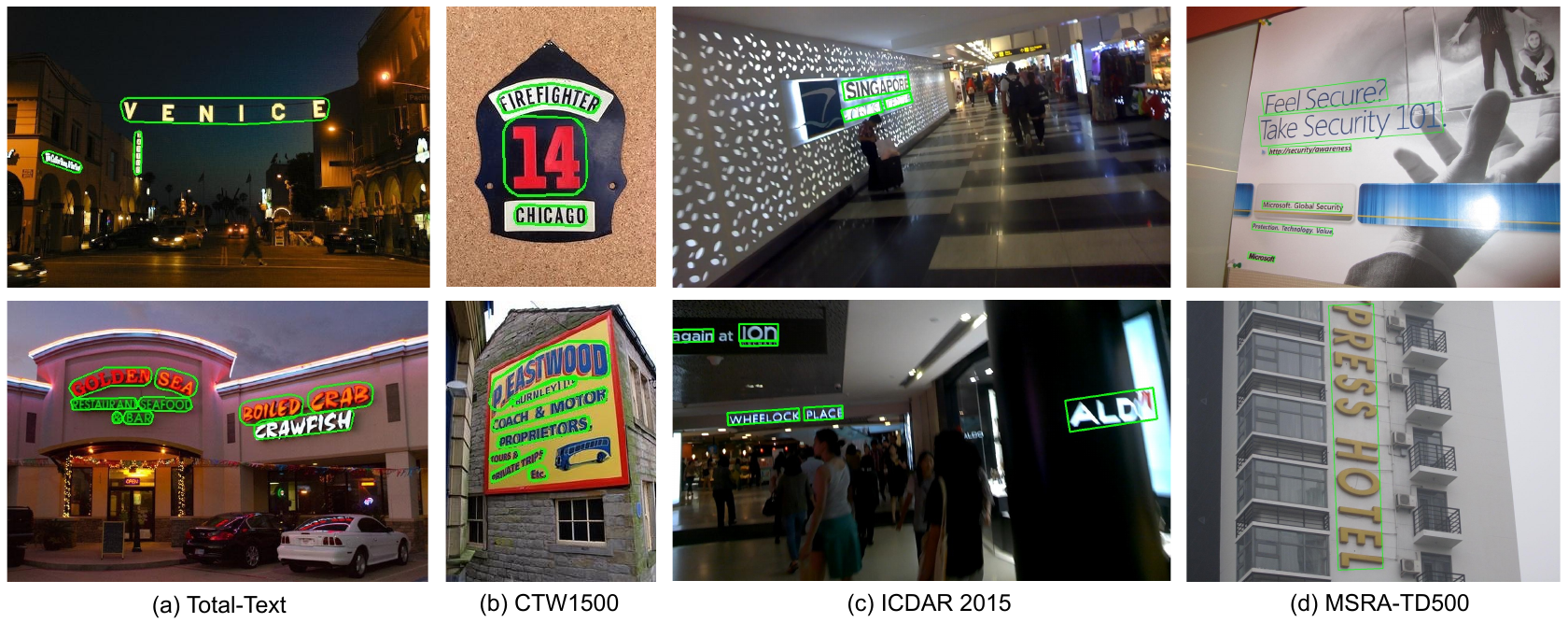}
\end{center}
\vspace{-10px}
\caption{Qualitative detection results of SIR on the Total-Text, CTW1500, ICDAR2015, and MSRA-TD500 datasets.}
\label{fig:vis}
\end{figure*}

%-------------------------------------------------------------------------
\subsection{Label Generation and Optimization}
\noindent\textbf{Label generation.}
Given an input training sample, we generate labels for the above-mentioned tasks with the raw annotations.

For the main semantic segmentation task, we assign a category label for each pixel to generate the ground truth mask. The categories include text kernel, text border \cite{wu2017self}, and non-text region. Text kernel is obtained by shrinking the text region in an instance-wise way following \cite{wang2019PSENet,wang2019PAN,liao2020db}. The region in the boundary between a text instance and the corresponding kernel is regarded as the text border. In default, the shrink ratio is set to 0.6 empirically.

For GDSC, the ground truth $T$ is generated in the same way as the semantic segmentation task. The only difference is $T$ is a binary mask that only considers text and non-text.

For TDM, the ground truth of the bounding box is obtained by finding the minimum and maximum values among the vertex coordinates of the polygon annotation. The instance label is set to the "text" category.

\noindent\textbf{Optimization.}
The network is optimized with multiple supervision and the overall loss $\mathcal{L}$ can be formulated as below:
\begin{equation}
\mathcal{L} = \mathcal{L}_{seg} + \lambda_{1}\mathcal{L}_{gdsc} + \lambda_{2}\mathcal{L}_{tdm},
\end{equation}
where $\mathcal{L}_{seg}$, $\mathcal{L}_{gdsc}$, and $\mathcal{L}_{tdm}$ denote the loss functions of the semantic segmentation, GDSC, and TDM. $\lambda_{1}$ and $\lambda_{2}$ are set to 3.0 and 1.0 respectively.

The loss of TDM contains multiple parts:
\begin{equation}
\mathcal{L}_{tdm} = \lambda_{3}\mathcal{L}_{rpn} + \lambda_{4}\mathcal{L}_{cls} + \mathcal{L}_{reg},
\end{equation}
where $\mathcal{L}_{rpn}$, $\mathcal{L}_{cls}$, and $\mathcal{L}_{reg}$ refer to the losses of anchor-based dense supervision and proposal-based sparse supervision, which are the same as that defined in Faster R-CNN \cite{ren2015fasterrcnn}. Both $\lambda_{3}$ and $\lambda_{4}$ are set to 1.0 empirically.

%------------------------------------------------------------------------
\section{Experiments}

\subsection{Datasets}
% \vspace{-5px}
\noindent\textbf{SynthText} \cite{gupta2016synthetic} is a synthetic dataset which consists of 800k images. These images are synthesized from 8k background images. This dataset is only used for pre-training when compared with the SOTA methods.

\noindent\textbf{ICDAR2015} \cite{karatzas2015icdar2015} is a multi-oriented text detection dataset only for English words, which includes 1000 training images and 500 testing images. The text regions are annotated with quadrilaterals. 

\noindent\textbf{MSRA-TD500} \cite{yao2012detecting} focus on the Chinese and English incidental text lines, which are always presented in a long and multi-oriented pattern.
Large variations of text scale and orientation are presented in this dataset. It consists of 300 training images and 200 testing images. Since the training images are too few, we follow the common practice of previous works \cite{zhou2017east,long2018textsnake,wang2019PAN} to add 400 images from HUST-TR400 \cite{yao2014unified} to the training data.

\noindent\textbf{Total-Text} \cite{ch2017total} has 1255 training images and 300 testing images, which contain horizontal, multi-oriented, and curve texts. Each text is labeled as a polygon at the word level.

\noindent\textbf{CTW1500} \cite{liu2019curved} contains 1000 training images and 500 testing images. There are 10751 text instances in total, where 3530 are curved texts. In this dataset, text instances are annotated with 14-point polygons at the text line level.

\subsection{Implementation Details}

\noindent\textbf{Training.} Pre-trained weights on ImageNet are used to initialize the backbone. The models are trained for 1200 epochs on the corresponding datasets with an SGD optimizer. The initial learning rate is set to 0.005 with a batch size of 16, following a ``poly'' learning rate policy as \cite{chen2017deeplab}. The warmup strategy is used in the first 1000 iterations for stable optimization. Additionally, when comparing with the SOTA methods, we first pre-train our models on the SynthText dataset for 2 epochs and then fine-tune the models on the corresponding datasets. 

Several data augmentation technologies are used in training: (1) Random flipping; (2) Random scaling of the image with a scale factor sampling from $[0.5, 3.0]$; (3) Random rotation with an angle range of $[-10^{\circ}, 10^{\circ}]$; (4) Random cropping and padding to the size of $640 \times 640$. 

\noindent\textbf{Inference.} During inference, the input image is resized along the short side with the respect ratio unchanged. The models are evaluated under the environment of a single GeForce RTX 2080 Ti GPU and an Intel Xeon Silver 4214R CPU with a batch size of 1.

%-------------------------------------------------------------------------
\subsection{Comparisons with State-of-the-Art Methods}
We compare the proposed method on four pubic datasets of different types including Total-Text, CTW1500, ICDAR2015, and MSRA-TD500 to demonstrate the effectiveness of SIR on real-time arbitrary-shaped scene text detection. As shown in Fig. \ref{fig:vis}, SIR is able to detect four types of scene text, which demonstrate robustness against large variations in aspect ratio and complex shape. Specially, we reproduce several real-time scene text detection methods including PAN\footnote{https://github.com/whai362/pan pp.pytorch.git} \cite{wang2019PAN} and DB/ DB++\footnote{https://github.com/MhLiao/DB.git} \cite{liao2020db,liao2022db++} from the corresponding official implementation for a fair comparison.

\begin{table*}[!htb]
\footnotesize
\centering
\caption{Detection results on the Total-Text, CTW1500, ICDAR2015, and MSRA-TD500 datasets. * denotes reproduced results by us. $\dagger$ denotes the use of character-level supervision.}
\vspace{-10px}
\begin{tabularx}{1.0\linewidth}{@{}lYYYYYYYYYYYYYYYYY@{}}
\toprule
\multirow{2}{*}{Method}                & \multirow{2}{*}{Ext} & \multicolumn{4}{c}{Total-Text}                       & \multicolumn{4}{c}{CTW1500}                                   & \multicolumn{4}{c}{ICDAR2015}                        & \multicolumn{4}{c}{MSRA-TD500}                                \\ \cline{3-18} 
                                       &                      & P             & R             & F             & FPS  & P             & R             & F             & FPS           & P             & R             & F             & FPS  & P             & R             & F             & FPS           \\ \midrule
TextField \cite{xu2019textfield}       & Synth                & 81.2          & 79.9          & 80.6          & -    & 83.0          & 79.8          & 81.4          & -             & 84.3          & 83.9          & 84.1          & -    & 87.4          & 75.9          & 81.3          & -             \\
LOMO \cite{zhang2019lomo}              & Synth                & 88.6          & 76.7          & 81.6          & 4.4  & \textbf{89.2} & 69.6          & 78.4          & 4.4           & 91.3          & 83.5          & 87.2          & 3.4  & -             & -             & -             & -             \\
CRAFT \cite{baek2019CRAFT}$\dagger$    & MLT17                & 87.6          & 79.9          & 83.6          & -    & 86.0          & 81.1          & 83.5          & -             & 89.8          & 84.3          & 86.9          & -    & 88.2          & 78.2          & 82.9          & -             \\
PSENet-1s   \cite{wang2019PSENet}      & MLT17                & 84.0          & 78.0          & 80.9          & 3.9  & 84.8          & 79.7          & 82.2          & 3.9           & 86.9          & 84.5          & 85.7          & 1.6  & -             & -             & -             & -             \\
ATRR \cite{wang2019ATRR}               & -                    & 80.9          & 76.2          & 78.5          & -    & 80.1          & 80.2          & 80.1          & -             & 89.2          & 86.0          & 87.6          & -    & 85.2          & 82.1          & 83.6          & -             \\
SAE    \cite{tian2019SAE}              & Synth                & -             & -             & -             & -    & 82.7          & 77.8          & 80.1          & -             & 88.3          & 85.0          & 86.6          & -    & 84.2          & 81.7          & 82.9          & -             \\
CSE \cite{liu2019CSE}                  & MLT17                & 81.4          & 79.1          & 80.2          & 2.4  & 81.1          & 76.0          & 78.4          & 2.6           & -             & -             & -             & -    & -             & -             & -             & -             \\
MSR \cite{xue2019msr}                  & Synth                & 83.8          & 74.8          & 79.0          & -    & 85.0          & 78.3          & 81.5          & 4.3           & 86.6          & 78.4          & 82.3          & 4.3  & 87.4          & 76.7          & 81.7          & -             \\
DRRG \cite{zhang2020DRRG}              & MLT17                & 86.5          & 84.9          & 85.7          & -    & 85.9          & 83.0          & 84.5          & -             & 88.5          & 84.7          & 86.6          & -    & 88.1          & 82.3          & 85.1          & -             \\
ContourNet   \cite{Wang2020contournet} & -                    & 86.9          & 83.9          & 85.4          & 3.8  & 83.7          & 84.1          & 83.9          & 4.5           & 87.6          & 86.1          & 86.9          & 3.5  & -             & -             & -             & -             \\
TextFuseNet   \cite{ye2020textfusenet}$\dagger$ & Synth                & 87.5          & 83.2          & 85.3          & 7.1  & 85.8          & \textbf{85.0} & 85.4          & 7.3           & 91.3          & \textbf{88.9} & \textbf{90.1} & 8.3  & -             & -             & -             & -             \\
SD \cite{xiao2020SD}$\dagger$          & MLT17                & 89.2          & 84.7          & 86.9          & -    & 85.8          & 82.3          & 84.0          & -             & 88.7          & 88.4          & 88.6          &      & -             & -             & -             & -             \\
TextRay \cite{Wang2020textray}         & ArT                  & 83.5          & 77.9          & 80.6          & -    & 82.8          & 80.4          & 81.6          & -             & -             & -             & -             & -    & -             & -             & -             & -             \\
FCENet \cite{zhu2021fce}               & -                    & 89.3          & 82.5          & 85.8          & -    & 87.6          & 83.4          & \textbf{85.5} & -             & 90.1          & 82.6          & 86.2          & -    & -             & -             & -             & -             \\
PCR \cite{dai2021pcr}                  & MLT17                & 88.5          & 82.0          & 85.2          & -    & 87.2          & 82.3          & 84.7          & -             & -             & -             & -             & -    & 90.8          & 83.5          & 87.0          & -             \\
BPN \cite{zhang2021textbpn}            & MLT17                & 90.7          & 85.2          & 87.9          & 10.7 & 86.5          & 83.6          & 85.0          & 12.2          & -             & -             & -             & -    & 86.6          & 84.5          & 85.6          & 12.3          \\
FSG  \cite{tang2022FSG}                & Synth                & 90.7          & \textbf{85.7} & 88.1          & -    & 88.1          & 82.4          & 85.2          & -             & 91.1          & 86.7          & 88.8          & -    & 91.4          & 84.7          & 87.9          & -             \\
PAN \cite{wang2019PAN}                 & Synth                & 89.3          & 81.0          & 85.0          & 39.6 & 86.4          & 81.2          & 83.7          & 39.8          & 84.0          & 81.9          & 82.9          & 26.1 & 84.4          & 83.8          & 84.1          & 30.2          \\
DB-R18 \cite{liao2020db}               & Synth                & 88.3          & 77.9          & 82.8          & 50.0 & 84.8          & 77.5          & 81.0          & 55.0          & 86.8          & 78.4          & 82.3          & 48.0 & 90.4          & 76.3          & 82.8          & 62.0          \\
DB-R50 \cite{liao2020db}               & Synth                & 87.1          & 82.5          & 84.7          & 32.0 & 86.9          & 80.2          & 83.4          & 22.0          & \textbf{91.8} & 83.2          & 87.3          & 12.0 & 91.5          & 79.2          & 84.9          & 32.0          \\
DB++-R18 \cite{liao2022db++}           & Synth                & 87.4          & 79.6          & 83.3          & 48.0 & 84.3          & 81.0          & 82.6          & 49.0          & 90.1          & 77.2          & 83.1          & 44,0 & 87.9          & 82.5          & 85.1          & 55.0          \\
DB++-R50 \cite{liao2022db++}           & Synth                & 88.9          & 83.2          & 86.0          & 28.0 & 87.9          & 82.8          & 85.3          & 26.0          & 90.9          & 83.9          & 87.3          & 10.0 & 91.5          & 83.3          & 87.2          & 29.0          \\ \midrule
PAN \cite{wang2019PAN}*                & Synth                & 88.5          & 81.7          & 85.0          & 41.3 & 86.0          & 81.0          & 83.4          & 40.4          & 86.6          & 79.7          & 83.0          & 31.5 & 85.7          & 83.4          & 84.5          & 43.3          \\
DB-R18 \cite{liao2020db}*              & Synth                & 89.7          & 78.2          & 83.5          & 49.3 & -             & -             & -             & -             & 87.6          & 77.5          & 82.2          & 44.1 & 86.8          & 79.9          & 83.2          & 55.7          \\
DB-R50 \cite{liao2020db}*              & Synth                & 87.6          & 81.2          & 84.3          & 26.4 & -             & -             & -             & -             & 90.9          & 84.2          & 87.4          & 10.2 & 91.7          & 81.1          & 86.1          & 29.4          \\
DB++-R18 \cite{liao2022db++}*          & Synth                & 89.0          & 78.0          & 83.1          & 44.1 & -             & -             & -             & -             & 90.0          & 77.2          & 83.1          & 40.1 & 89.6          & 81.3          & 85.2          & 48.8          \\
DB++-R50 \cite{liao2022db++}*          & Synth                & 89.2          & 82.8          & 85.9          & 24.8 & -             & -             & -             & -             & 90.7          & 83.8          & 87.1          & 9.7  & 92.4          & 83.2          & 87.5          & 27.7          \\ \midrule
SIR-R18 (Ours)                         & Synth                & \textbf{91.8} & 83.1          & 87.2          & 48.2 & 84.7          & 82.9          & 83.8          & \textbf{58.8} & 89.2          & 80.7          & 84.7          & 38.8 & 89.4          & 83.6          & 86.4          & \textbf{58.0} \\
SIR-R50 (Ours)                         & Synth                & 90.9          & 85.6          & \textbf{88.2} & 30.9 & 87.4          & 83.7          & \textbf{85.5} & 38.7          & 90.4          & 85.4          & 87.8          & 10.1 & \textbf{93.6} & \textbf{86.0} & \textbf{89.6} & 36.9          \\ \bottomrule
\end{tabularx}
\label{tab:sota}
\end{table*}

\noindent\textbf{Detecting curved words.} We verify the ability of our method to detect curved words on Total-Text. The test scale is set to 640. As listed in Table \ref{tab:sota}, on Total-Text, SIR-R50 achieves the SOTA performance of 88.2\% F-measure with 30.9 FPS which outperforms the recent SOTA methods BPN \cite{zhang2021textbpn} and FSG \cite{tang2022FSG} and runs much faster. Compared with BPN \cite{zhang2021textbpn} and FSG \cite{tang2022FSG} which involves several refinement stages, SIR directly learns robust representation with auxiliary tasks without burdening the inference speed. SIR could outperform DB++ \cite{liao2022db++} on both accuracy and speed. With ResNet-18, SIR achieves 87.2\% F-measure with 48.2 FPS, which is only a little slower than DB \cite{liao2020db} with 49.3 FPS, but outperforms DB by a large margin of 3.7\% in terms of F-measure, demonstrating the superiority of SIR. 

\noindent\textbf{Detecting curved lines.}
We verify the ability of the proposed method to detect curved lines on CTW1500. The short side is set to 512. As listed in Table \ref{tab:sota}, on CTW1500, a competitive result of 85.5\% with 38.7 FPS is achieved by SIR-R50, which share the same performance with the SOTA methods FCENet \cite{zhu2021fce} and outperform recently proposed FSG \cite{tang2022FSG} and BPN \cite{zhang2021textbpn}. With ResNet-18, SIR archives 83.8\% F-measure with 58.8 FPS, which is faster and better in terms of FPS and F-measure than PAN \cite{wang2019PAN}, DB \cite{liao2020db}, and DB++ \cite{liao2022db++}, which illustrate the ability on detecting long curve text.

\noindent\textbf{Detecting oriented words.}
The test scale on ICDAR2015 is set to 736 for SIR-R18 and 1152 for SIR-R50 following \cite{liao2022db++}. As shown in Table \ref{tab:sota}, SIR achieves a competitive performance of $87.8\%$ F-measure, which is comparable to FSG \cite{tang2022FSG} and SD \cite{xiao2020SD} but inferior to TextFuseNet \cite{ye2020textfusenet}. However, TextFuseNet uses character-level supervision, which is not suitable for direct comparison.
SIR also outperforms other real-time works, e.g., PAN \cite{wang2019PAN}, DB \cite{liao2020db}, and DB++ \cite{liao2022db++}. With ResNet-18, SIR gets 84.7\% F-measure with 38.8 FPS, which is faster and better than PAN with 83.0 F-measure and 31.5 FPS. SIR runs as fast as DB++ but outperforms it by 1.6\%, which illustrates the ability of SIR in challenging blur text situations.

\noindent\textbf{Detecting oriented lines.}
The long text is the common case in the natural scene. We test the robustness of our models of detection of long text lines on MSRA-TD500. The test scale is 640. As shown in Table \ref{tab:sota}, our proposed method shows powerful ability in long text detection. SIR-R50 achieves 89.6\% F-measure with 36.9 FPS, outperforms all the previous methods, and boosts the best performance on this dataset by $1.7\%$ in terms of F-measure, which demonstrates that SIR can deal with large variations in scales and rotation well. Note that even with a light backbone ResNet-18, SIR achieves 86.4 F-measure and the fastest speed of 58.0 FPS, already outperforming most existing methods, which illustrates the ability of SIR on learning discriminative features.

\subsection{Ablation Study}

The ablation study is conducted on an oriented text dataset ICDAR2015 and a curved text dataset Total-Text to demonstrate the effectiveness of GDSC and TDM. ResNet-18 is used as the backbone. The test scale is also set to 736 and 640 on the two datasets.

\noindent\textbf{The effectiveness of GDSC/TDM.}
As shown in Table \ref{tab:ablation}, we can infer that both GDSC and TDM improve the performance of the baseline model on the two datasets, even though the BU-SEG baseline is quite strong. GDSC achieves 1.6\% and 1.3\% performance gain in terms of F-measure on the ICDAR2015 dataset and the Total-Text dataset, which illustrate the effectiveness of GDSC. TDM brings 1.8\% (on ICDAR2015) and 2.1\% (on Total-Text) improvements, demonstrating the superiority of TDM.

\noindent\textbf{The complementarity between GDSC and TDM.}
As shown in Table \ref{tab:ablation}, due to the modeling from different perspectives, GDSC and TDM act in a complementary manner. Specifically, when combining TDM, the model with GDSC is further improved by 1.2\% (on the ICDAR2015 dataset) and 1.5\% (on the Total-Text dataset); With GDSC, the baseline with TDM is increased by 1.0\% and 0.7\% on the ICDAR2015 dataset and the Total-Text dataset respectively.

\begin{table}[!htb]
\caption{Detection results with different settings of GDSC and TDM. ``Seg'' denotes the segmentation baseline. }
\vspace{-10px}
\centering
\footnotesize
\begin{tabularx}{1.0\linewidth}{@{}YYYYYYYYY@{}}
\toprule
\multirow{2}{*}{Seg} & \multirow{2}{*}{GDSC} & \multirow{2}{*}{TDM} & \multicolumn{3}{c}{ICDAR2015} & \multicolumn{3}{c}{Total-Text} \\ \cline{4-9} 
                     &                       &                      & P        & R        & F       & P        & R        & F        \\ \midrule
\checkmark           & $\times$              & $\times$             & 85.0     & 77.8     & 81.2    & 88.8     & 79.2     & 83.7     \\
\checkmark           & \checkmark            & $\times$             & 87.9     & 78.2     & 82.8    & 87.0     & 83.2     & 85.0     \\
\checkmark           & $\times$              & \checkmark           & 84.6     & \textbf{81.5}     & 83.0      & 89.4     & 82.5     & 85.8     \\
\checkmark           & \checkmark            & \checkmark           & \textbf{88.7}  & 79.7 & \textbf{84.0} &  88.2  & \textbf{84.8} & \textbf{86.5}\\ \bottomrule
\end{tabularx}
\label{tab:ablation}
\end{table}

\begin{figure}[!htb]
\begin{center}
\includegraphics[width=0.8\linewidth]{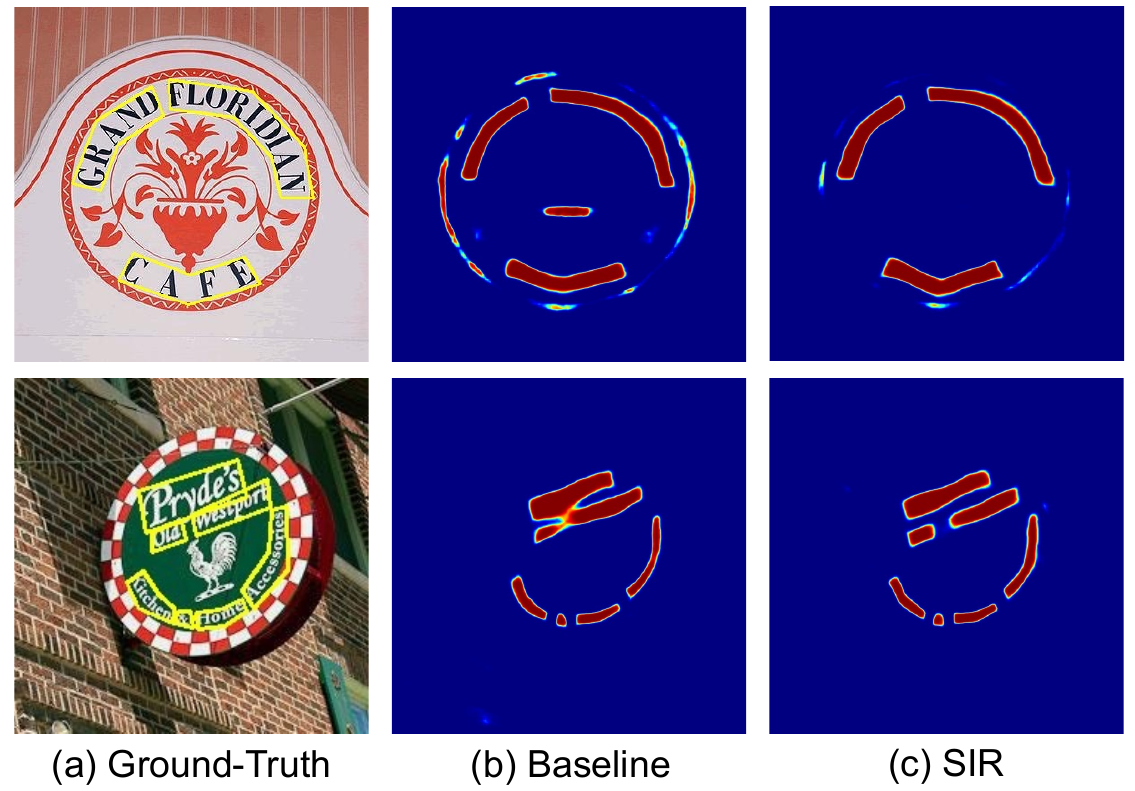}
\end{center}
\vspace{-10px}
\caption{Visualization of the probability maps generated by the baseline segmentation model (b) and SIR (c). (a) denotes the ground truth.}
\label{fig:ablation}
\end{figure}

\noindent\textbf{Comparison between the baseline and SIR.} In Table \ref{tab:ablation}, with GDSC and TDM, SIR outperforms the baseline model by 2.8\% on both the ICDAR2015 dataset and the Total-Text dataset quantitatively in total. We further provide qualitative results for better illustration. As shown in Fig. \ref{fig:ablation}, the baseline model tends to suffer from false positives and text adhesion issues, which alleviate the robustness of the model. Due to auxiliary tasks of GDSC and TDM, which promote the representation learning on semantic and instance level, SIR learns robust representation and produce high-quality probability maps as well as accurate detection.

\noindent\textbf{The function of modules in TDM.} We perform a detailed study on the modules of TDM. The result is listed in Table \ref{tab:TDM}. Several conclusions can be drawn:
\textbf{(1) Effectiveness of dense anchor-based supervision.} RPN brings 0.5\% and 1.2\% improvements in terms of F-measure on the ICDAR2015 dataset and the Total-Text dataset. 
\textbf{(2) Both regional classification and regression help.} When further combining ``Cls'', the model achieves 0.6\% and 0.7\% performance gain on the ICDAR2015 dataset and the Total-Text dataset; ``Reg'' brings 0.8\% and 0.7\% improvements on the ICDAR2015 dataset and the Total-Text dataset.
\textbf{(3) Regional recognition tasks can be performed even with anchors.} As shown in Table \ref{tab:TDM}, when RPN is removed in which the anchors are directly taken as proposals, regional recognition tasks can bring 0.5\% and 1.2\% improvements on the ICDAR2015 dataset and the Total-Text dataset.

\begin{table}[!htb]
\caption{Detection results with different settings of TDM. ``Cls'' and ``Reg'' refer to the classification and regression tasks in regional recognition.}
\vspace{-10px}
\centering
\footnotesize
\begin{tabularx}{1.0\linewidth}{@{}YYYYYYYYYY@{}}
\toprule
\multirow{2}{*}{Seg} & \multirow{2}{*}{RPN} & \multirow{2}{*}{Cls} & \multirow{2}{*}{Reg} & \multicolumn{3}{c}{ICDAR2015} & \multicolumn{3}{c}{Total-Text} \\ \cline{5-10} 
                     &                      &                      &                      & P        & R        & F       & P        & R        & F        \\ \midrule 
\checkmark           & $\times$             & $\times$             & $\times$             & 85.0     & 77.8     & 81.2    & 88.8     & 79.2     & 83.7     \\
\checkmark           & \checkmark           & $\times$             & $\times$             & 85.2     & 78.5     & 81.7    & 89.7     & 80.6     & 84.9     \\
\checkmark           & \checkmark           & \checkmark           & $\times$             & \textbf{86.9} & 78.2     & 82.3    & \textbf{89.9} & 81.8     & 85.6     \\
\checkmark           & \checkmark           & $\times$             & \checkmark           & 85.9     & 79.3     & 82.5    & 86.9     & \textbf{84.3}     & 85.6     \\
\checkmark           & \checkmark           & \checkmark           & \checkmark           & 84.6     & \textbf{81.5} & \textbf{83.0} & 89.4     & 82.5     & \textbf{85.8}\\
\checkmark           & $\times$             & \checkmark           & \checkmark           & 84.4     & 79.1     & 81.7    & 89.7     & 80.5     & 84.9     \\ \bottomrule
\end{tabularx}
\label{tab:TDM}
\end{table}

\noindent\textbf{Evaluation of inference time and training time.}  
As shown in Table \ref{tab:speed}, compared to the baseline model, SIR not only brings no extra burden but also runs 
a little faster than the baseline model during inference. We find it comes from the reduction of the number of predicted boxes in the post-process, which also demonstrates the improvement in the quality of the probability map through robust representation learning.
During training, SIR acts like a two-stage detector, compared to the one-stage baseline model, which can be seen as Faster R-CNN compared to RPN during training. SIR costs 0.72 seconds per iteration against 0.47 seconds per iteration for the baseline model, which brings about 50\% extra training time and is acceptable in real applications due to the efficiency of the baseline.

\begin{table}[!htb]
\caption{Comparison of the number of processed boxes and the running time between the baseline and SIR.}
\vspace{-10px}
\centering
\footnotesize
\begin{tabularx}{1.0\linewidth}{@{}lYYYY@{}}
\toprule
\multirow{2}{*}{Method} & \multicolumn{2}{c}{Box number (per image)} & \multicolumn{2}{c}{Inference speed (FPS)} \\ \cline{2-5} 
                        & ICDAR2015        & Total-Text       & ICDAR2015           & Total-Text          \\ \hline
Baseline                & 12.4             & 10.8             & 38.1                & 47.4                \\
SIR (Ours)              & 10.8             & 9.8              & 38.8                & 48.2                \\ \hline
\end{tabularx}
\label{tab:speed}
\end{table}
\vspace{-10px}

\subsection{Limitation}
Our method shares the same limitation with existing BU-SEG methods \cite{liao2022db++,liao2020db,wang2019PAN,wang2019PSENet}. Text instances with a large overlap in the text kernel will lead to a single connected component, which can not be well detected by the proposed method.

%-------------------------------------------------------------------------
\section{Conclusion}
% \vspace{-5px}
In this paper, we propose a robust real-time scene text detection method (SIR) to improve the robustness of the existing encoder-decoder segmentation-based framework. SIR originates from the perspective of representation learning and consists of a unified BU-SEG baseline, GDSC, and TDM in which GDSC and TDM act as auxiliary tasks that only work in training and are removed during inference. GDSC learns the semantic representation by contrasting a global semantic representation to a dense representation. TDM performs regional recognition tasks for instance-level representation learning that bring strong instance clues to the encoder. As a result, robust representations are learned by the encoder, which effectively mitigates the text/non-text pixel misclassification and false text instance discrimination issues in the existing encoder-decoder framework. Experimental results on four public datasets demonstrate the efficiency and robustness of the proposed method. 
\begin{acks}
Supported by the Key Research Program of Frontier Sciences, CAS, Grant NO ZDBS-LY-7024.
\end{acks}

\bibliographystyle{ACM-Reference-Format}
% \balance
\bibliography{acmart}


\begin{thebibliography}{100}

%%% the Web address.

\ifx \showCODEN    \undefined \def \showCODEN     #1{\unskip}     \fi
\ifx \showDOI      \undefined \def \showDOI       #1{#1}\fi
\ifx \showISBNx    \undefined \def \showISBNx     #1{\unskip}     \fi
\ifx \showISBNxiii \undefined \def \showISBNxiii  #1{\unskip}     \fi
\ifx \showISSN     \undefined \def \showISSN      #1{\unskip}     \fi
\ifx \showLCCN     \undefined \def \showLCCN      #1{\unskip}     \fi
\ifx \shownote     \undefined \def \shownote      #1{#1}          \fi
\ifx \showarticletitle \undefined \def \showarticletitle #1{#1}   \fi
\ifx \showURL      \undefined \def \showURL       {\relax}        \fi
% invisible to TeX
\providecommand\bibfield[2]{#2}
\providecommand\bibinfo[2]{#2}
\providecommand\natexlab[1]{#1}
\providecommand\showeprint[2][]{arXiv:#2}

\bibitem[Baek et~al\mbox{.}(2019)]%
        {baek2019CRAFT}
\bibfield{author}{\bibinfo{person}{Youngmin Baek}, \bibinfo{person}{Bado Lee},
  \bibinfo{person}{Dongyoon Han}, \bibinfo{person}{Sangdoo Yun}, {and}
  \bibinfo{person}{Hwalsuk Lee}.} \bibinfo{year}{2019}\natexlab{}.
\newblock \showarticletitle{Character region awareness for text detection}. In
  \bibinfo{booktitle}{\emph{CVPR}}. \bibinfo{pages}{9365--9374}.
\newblock


\bibitem[Chen et~al\mbox{.}(2017)]%
        {chen2017deeplab}
\bibfield{author}{\bibinfo{person}{Liang-Chieh Chen}, \bibinfo{person}{George
  Papandreou}, \bibinfo{person}{Iasonas Kokkinos}, \bibinfo{person}{Kevin
  Murphy}, {and} \bibinfo{person}{Alan~L Yuille}.}
  \bibinfo{year}{2017}\natexlab{}.
\newblock \showarticletitle{Deeplab: Semantic image segmentation with deep
  convolutional nets, atrous convolution, and fully connected crfs}.
\newblock \bibinfo{journal}{\emph{IEEE TPAMI}} \bibinfo{volume}{40},
  \bibinfo{number}{4} (\bibinfo{year}{2017}), \bibinfo{pages}{834--848}.
\newblock


\bibitem[Ch'ng and Chan(2017)]%
        {ch2017total}
\bibfield{author}{\bibinfo{person}{Chee~Kheng Ch'ng} {and}
  \bibinfo{person}{Chee~Seng Chan}.} \bibinfo{year}{2017}\natexlab{}.
\newblock \showarticletitle{Total-text: A comprehensive dataset for scene text
  detection and recognition}. In \bibinfo{booktitle}{\emph{ICDAR}}.
  \bibinfo{pages}{935--942}.
\newblock


\bibitem[Dai et~al\mbox{.}(2021)]%
        {dai2021pcr}
\bibfield{author}{\bibinfo{person}{Pengwen Dai}, \bibinfo{person}{Sanyi Zhang},
  \bibinfo{person}{Hua Zhang}, {and} \bibinfo{person}{Xiaochun Cao}.}
  \bibinfo{year}{2021}\natexlab{}.
\newblock \showarticletitle{Progressive Contour Regression for Arbitrary-Shape
  Scene Text Detection}. In \bibinfo{booktitle}{\emph{CVPR}}.
  \bibinfo{pages}{7393--7402}.
\newblock


\bibitem[Deng et~al\mbox{.}(2018)]%
        {deng2018pixellink}
\bibfield{author}{\bibinfo{person}{Dan Deng}, \bibinfo{person}{Haifeng Liu},
  \bibinfo{person}{Xuelong Li}, {and} \bibinfo{person}{Deng Cai}.}
  \bibinfo{year}{2018}\natexlab{}.
\newblock \showarticletitle{Pixellink: Detecting scene text via instance
  segmentation}. In \bibinfo{booktitle}{\emph{AAAI}}.
  \bibinfo{pages}{6773--6780}.
\newblock


\bibitem[Du et~al\mbox{.}(2022)]%
        {du2022svtr}
\bibfield{author}{\bibinfo{person}{Yongkun Du}, \bibinfo{person}{Zhineng Chen},
  \bibinfo{person}{Caiyan Jia}, \bibinfo{person}{Xiaoting Yin},
  \bibinfo{person}{Tianlun Zheng}, \bibinfo{person}{Chenxia Li},
  \bibinfo{person}{Yuning Du}, {and} \bibinfo{person}{Yu{-}Gang Jiang}.}
  \bibinfo{year}{2022}\natexlab{}.
\newblock \showarticletitle{{SVTR:} Scene Text Recognition with a Single Visual
  Model}. In \bibinfo{booktitle}{\emph{IJCAI}}. \bibinfo{pages}{884--890}.
\newblock


\bibitem[Guo et~al\mbox{.}(2021)]%
        {guo2021icann}
\bibfield{author}{\bibinfo{person}{Youhui Guo}, \bibinfo{person}{Yu Zhou},
  \bibinfo{person}{Xugong Qin}, {and} \bibinfo{person}{Weiping Wang}.}
  \bibinfo{year}{2021}\natexlab{}.
\newblock \showarticletitle{Which and Where to Focus: {A} Simple yet Accurate
  Framework for Arbitrary-Shaped Nearby Text Detection in Scene Images}. In
  \bibinfo{booktitle}{\emph{ICANN}}. \bibinfo{pages}{271--283}.
\newblock


\bibitem[Guo et~al\mbox{.}(2022)]%
        {guo2022icme}
\bibfield{author}{\bibinfo{person}{Youhui Guo}, \bibinfo{person}{Yu Zhou},
  \bibinfo{person}{Xugong Qin}, \bibinfo{person}{Enze Xie}, {and}
  \bibinfo{person}{Weiping Wang}.} \bibinfo{year}{2022}\natexlab{}.
\newblock \showarticletitle{{UNITS:} Unsupervised Intermediate Training Stage
  for Scene Text Detection}. In \bibinfo{booktitle}{\emph{ICME}}.
  \bibinfo{pages}{1--6}.
\newblock


\bibitem[Gupta et~al\mbox{.}(2016)]%
        {gupta2016synthetic}
\bibfield{author}{\bibinfo{person}{Ankush Gupta}, \bibinfo{person}{Andrea
  Vedaldi}, {and} \bibinfo{person}{Andrew Zisserman}.}
  \bibinfo{year}{2016}\natexlab{}.
\newblock \showarticletitle{Synthetic data for text localisation in natural
  images}. In \bibinfo{booktitle}{\emph{CVPR}}. \bibinfo{pages}{2315--2324}.
\newblock


\bibitem[He et~al\mbox{.}(2017c)]%
        {he2017cias}
\bibfield{author}{\bibinfo{person}{Dafang He}, \bibinfo{person}{Xiao Yang},
  \bibinfo{person}{Chen Liang}, \bibinfo{person}{Zihan Zhou},
  \bibinfo{person}{Alexander~G Ororbi}, \bibinfo{person}{Daniel Kifer}, {and}
  \bibinfo{person}{C Lee~Giles}.} \bibinfo{year}{2017}\natexlab{c}.
\newblock \showarticletitle{Multi-scale FCN with cascaded instance aware
  segmentation for arbitrary oriented word spotting in the wild}. In
  \bibinfo{booktitle}{\emph{CVPR}}. \bibinfo{pages}{3519--3528}.
\newblock


\bibitem[He et~al\mbox{.}(2017a)]%
        {he2017maskrcnn}
\bibfield{author}{\bibinfo{person}{Kaiming He}, \bibinfo{person}{Georgia
  Gkioxari}, \bibinfo{person}{Piotr Doll{\'a}r}, {and} \bibinfo{person}{Ross
  Girshick}.} \bibinfo{year}{2017}\natexlab{a}.
\newblock \showarticletitle{Mask R-CNN}. In \bibinfo{booktitle}{\emph{ICCV}}.
  \bibinfo{pages}{2980--2988}.
\newblock


\bibitem[He et~al\mbox{.}(2016)]%
        {he2016resnet}
\bibfield{author}{\bibinfo{person}{Kaiming He}, \bibinfo{person}{Xiangyu
  Zhang}, \bibinfo{person}{Shaoqing Ren}, {and} \bibinfo{person}{Jian Sun}.}
  \bibinfo{year}{2016}\natexlab{}.
\newblock \showarticletitle{Deep residual learning for image recognition}. In
  \bibinfo{booktitle}{\emph{CVPR}}. \bibinfo{pages}{770--778}.
\newblock


\bibitem[He et~al\mbox{.}(2021)]%
        {he2021most}
\bibfield{author}{\bibinfo{person}{Minghang He}, \bibinfo{person}{Minghui
  Liao}, \bibinfo{person}{Zhibo Yang}, \bibinfo{person}{Humen Zhong},
  \bibinfo{person}{Jun Tang}, \bibinfo{person}{Wenqing Cheng},
  \bibinfo{person}{Cong Yao}, \bibinfo{person}{Yongpan Wang}, {and}
  \bibinfo{person}{Xiang Bai}.} \bibinfo{year}{2021}\natexlab{}.
\newblock \showarticletitle{MOST: A Multi-Oriented Scene Text Detector with
  Localization Refinement}. In \bibinfo{booktitle}{\emph{CVPR}}.
  \bibinfo{pages}{8813--8822}.
\newblock


\bibitem[He et~al\mbox{.}(2017b)]%
        {he2017SSTD}
\bibfield{author}{\bibinfo{person}{Pan He}, \bibinfo{person}{Weilin Huang},
  \bibinfo{person}{Tong He}, \bibinfo{person}{Qile Zhu}, \bibinfo{person}{Yu
  Qiao}, {and} \bibinfo{person}{Xiaolin Li}.} \bibinfo{year}{2017}\natexlab{b}.
\newblock \showarticletitle{Single shot text detector with regional attention}.
  In \bibinfo{booktitle}{\emph{ICCV}}. \bibinfo{pages}{3047--3055}.
\newblock


\bibitem[He et~al\mbox{.}(2017d)]%
        {he2017deepregression}
\bibfield{author}{\bibinfo{person}{Wenhao He}, \bibinfo{person}{Xu-Yao Zhang},
  \bibinfo{person}{Fei Yin}, {and} \bibinfo{person}{Cheng-Lin Liu}.}
  \bibinfo{year}{2017}\natexlab{d}.
\newblock \showarticletitle{Deep direct regression for multi-oriented scene
  text detection}. In \bibinfo{booktitle}{\emph{ICCV}}.
  \bibinfo{pages}{745--753}.
\newblock


\bibitem[Hu et~al\mbox{.}(2020)]%
        {hu2020gtc}
\bibfield{author}{\bibinfo{person}{Wenyang Hu}, \bibinfo{person}{Xiaocong Cai},
  \bibinfo{person}{Jun Hou}, \bibinfo{person}{Shuai Yi}, {and}
  \bibinfo{person}{Zhiping Lin}.} \bibinfo{year}{2020}\natexlab{}.
\newblock \showarticletitle{Gtc: Guided training of ctc towards efficient and
  accurate scene text recognition}. In \bibinfo{booktitle}{\emph{AAAI}}.
  \bibinfo{pages}{11005--11012}.
\newblock


\bibitem[Huang et~al\mbox{.}(2014)]%
        {huang2014robust}
\bibfield{author}{\bibinfo{person}{Weilin Huang}, \bibinfo{person}{Yu Qiao},
  {and} \bibinfo{person}{Xiaoou Tang}.} \bibinfo{year}{2014}\natexlab{}.
\newblock \showarticletitle{Robust scene text detection with convolution neural
  network induced mser trees}. In \bibinfo{booktitle}{\emph{ECCV}}.
  \bibinfo{pages}{497--511}.
\newblock


\bibitem[Jia et~al\mbox{.}(2016)]%
        {jia2016DFN}
\bibfield{author}{\bibinfo{person}{Xu Jia}, \bibinfo{person}{Bert
  De~Brabandere}, \bibinfo{person}{Tinne Tuytelaars}, {and}
  \bibinfo{person}{Luc~V Gool}.} \bibinfo{year}{2016}\natexlab{}.
\newblock \showarticletitle{Dynamic filter networks}. In
  \bibinfo{booktitle}{\emph{NeurIPS}}. \bibinfo{pages}{667--675}.
\newblock


\bibitem[Karatzas et~al\mbox{.}(2015)]%
        {karatzas2015icdar2015}
\bibfield{author}{\bibinfo{person}{Dimosthenis Karatzas},
  \bibinfo{person}{Lluis Gomez-Bigorda}, \bibinfo{person}{Anguelos Nicolaou},
  \bibinfo{person}{Suman Ghosh}, \bibinfo{person}{Andrew Bagdanov},
  \bibinfo{person}{Masakazu Iwamura}, \bibinfo{person}{Jiri Matas},
  \bibinfo{person}{Lukas Neumann}, \bibinfo{person}{Vijay~Ramaseshan
  Chandrasekhar}, \bibinfo{person}{Shijian Lu}, {et~al\mbox{.}}}
  \bibinfo{year}{2015}\natexlab{}.
\newblock \showarticletitle{ICDAR 2015 competition on robust reading}. In
  \bibinfo{booktitle}{\emph{ICDAR}}. \bibinfo{pages}{1156--1160}.
\newblock


\bibitem[Liao et~al\mbox{.}(2021)]%
        {liao2019masktextspotterv2}
\bibfield{author}{\bibinfo{person}{Minghui Liao}, \bibinfo{person}{Pengyuan
  Lyu}, \bibinfo{person}{Minghang He}, \bibinfo{person}{Cong Yao},
  \bibinfo{person}{Wenhao Wu}, {and} \bibinfo{person}{Xiang Bai}.}
  \bibinfo{year}{2021}\natexlab{}.
\newblock \showarticletitle{Mask textSpotter: An end-to-end trainable neural
  network for spotting text with arbitrary shapes}.
\newblock \bibinfo{journal}{\emph{IEEE TPAMI}} (\bibinfo{year}{2021}),
  \bibinfo{pages}{532--548}.
\newblock


\bibitem[Liao et~al\mbox{.}(2018a)]%
        {liao2018textboxes++}
\bibfield{author}{\bibinfo{person}{Minghui Liao}, \bibinfo{person}{Baoguang
  Shi}, {and} \bibinfo{person}{Xiang Bai}.} \bibinfo{year}{2018}\natexlab{a}.
\newblock \showarticletitle{Textboxes++: A single-shot oriented scene text
  detector}.
\newblock \bibinfo{journal}{\emph{IEEE TIP}} (\bibinfo{year}{2018}),
  \bibinfo{pages}{3676--3690}.
\newblock


\bibitem[Liao et~al\mbox{.}(2020)]%
        {liao2020db}
\bibfield{author}{\bibinfo{person}{Minghui Liao}, \bibinfo{person}{Zhaoyi Wan},
  \bibinfo{person}{Cong Yao}, \bibinfo{person}{Kai Chen}, {and}
  \bibinfo{person}{Xiang Bai}.} \bibinfo{year}{2020}\natexlab{}.
\newblock \showarticletitle{Real-time scene text detection with differentiable
  binarization}. In \bibinfo{booktitle}{\emph{AAAI}}.
  \bibinfo{pages}{11474--11481}.
\newblock


\bibitem[Liao et~al\mbox{.}(2018b)]%
        {liao2018RRD}
\bibfield{author}{\bibinfo{person}{Minghui Liao}, \bibinfo{person}{Zhen Zhu},
  \bibinfo{person}{Baoguang Shi}, \bibinfo{person}{Gui-song Xia}, {and}
  \bibinfo{person}{Xiang Bai}.} \bibinfo{year}{2018}\natexlab{b}.
\newblock In \bibinfo{booktitle}{\emph{CVPR}}. \bibinfo{pages}{5909--5918}.
\newblock


\bibitem[Liao et~al\mbox{.}(2023)]%
        {liao2022db++}
\bibfield{author}{\bibinfo{person}{Minghui Liao}, \bibinfo{person}{Zhisheng
  Zou}, \bibinfo{person}{Zhaoyi Wan}, \bibinfo{person}{Cong Yao}, {and}
  \bibinfo{person}{Xiang Bai}.} \bibinfo{year}{2023}\natexlab{}.
\newblock \showarticletitle{Real-time scene text detection with differentiable
  binarization and adaptive scale fusion}.
\newblock \bibinfo{journal}{\emph{IEEE TPAMI}} \bibinfo{volume}{45},
  \bibinfo{number}{1} (\bibinfo{year}{2023}), \bibinfo{pages}{919--931}.
\newblock


\bibitem[Lin et~al\mbox{.}(2017)]%
        {lin2017FPN}
\bibfield{author}{\bibinfo{person}{Tsung-Yi Lin}, \bibinfo{person}{Piotr
  Doll{\'a}r}, \bibinfo{person}{Ross Girshick}, \bibinfo{person}{Kaiming He},
  \bibinfo{person}{Bharath Hariharan}, {and} \bibinfo{person}{Serge Belongie}.}
  \bibinfo{year}{2017}\natexlab{}.
\newblock \showarticletitle{Feature pyramid networks for object detection}. In
  \bibinfo{booktitle}{\emph{CVPR}}. \bibinfo{pages}{2117--2125}.
\newblock


\bibitem[Liu et~al\mbox{.}(2016)]%
        {liu2016ssd}
\bibfield{author}{\bibinfo{person}{Wei Liu}, \bibinfo{person}{Dragomir
  Anguelov}, \bibinfo{person}{Dumitru Erhan}, \bibinfo{person}{Christian
  Szegedy}, \bibinfo{person}{Scott Reed}, \bibinfo{person}{Cheng-Yang Fu},
  {and} \bibinfo{person}{Alexander~C Berg}.} \bibinfo{year}{2016}\natexlab{}.
\newblock \showarticletitle{SSD: Single shot multibox detector}. In
  \bibinfo{booktitle}{\emph{ECCV}}. \bibinfo{pages}{21--37}.
\newblock


\bibitem[Liu et~al\mbox{.}(2020a)]%
        {Liu2020ABCNet}
\bibfield{author}{\bibinfo{person}{Yuliang Liu}, \bibinfo{person}{Hao Chen},
  \bibinfo{person}{Chunhua Shen}, \bibinfo{person}{Tong He},
  \bibinfo{person}{Lianwen Jin}, {and} \bibinfo{person}{Liangwei Wang}.}
  \bibinfo{year}{2020}\natexlab{a}.
\newblock \showarticletitle{ABCNet: Real-time scene text spotting with adaptive
  bezier-curve network}. In \bibinfo{booktitle}{\emph{CVPR}}.
  \bibinfo{pages}{9806--9815}.
\newblock


\bibitem[Liu and Jin(2017)]%
        {liu2017DMPNet}
\bibfield{author}{\bibinfo{person}{Yuliang Liu} {and} \bibinfo{person}{Lianwen
  Jin}.} \bibinfo{year}{2017}\natexlab{}.
\newblock \showarticletitle{Deep matching prior network: Toward tighter
  multi-oriented text detection}. In \bibinfo{booktitle}{\emph{CVPR}}.
  \bibinfo{pages}{3454--3461}.
\newblock


\bibitem[Liu et~al\mbox{.}(2020b)]%
        {liu2019maskTTD}
\bibfield{author}{\bibinfo{person}{Yuliang Liu}, \bibinfo{person}{Lianwen Jin},
  {and} \bibinfo{person}{Chuanming Fang}.} \bibinfo{year}{2020}\natexlab{b}.
\newblock \showarticletitle{Arbitrarily shaped scene text detection with a mask
  tightness text detector}.
\newblock \bibinfo{journal}{\emph{IEEE TIP}} (\bibinfo{year}{2020}),
  \bibinfo{pages}{2918--2930}.
\newblock


\bibitem[Liu et~al\mbox{.}(2019a)]%
        {liu2019curved}
\bibfield{author}{\bibinfo{person}{Yuliang Liu}, \bibinfo{person}{Lianwen Jin},
  \bibinfo{person}{Shuaitao Zhang}, \bibinfo{person}{Canjie Luo}, {and}
  \bibinfo{person}{Sheng Zhang}.} \bibinfo{year}{2019}\natexlab{a}.
\newblock \showarticletitle{Curved scene text detection via transverse and
  longitudinal sequence connection}.
\newblock \bibinfo{journal}{\emph{PR}} (\bibinfo{year}{2019}),
  \bibinfo{pages}{337--345}.
\newblock


\bibitem[Liu et~al\mbox{.}(2019c)]%
        {liu2019BDN}
\bibfield{author}{\bibinfo{person}{Yuliang Liu}, \bibinfo{person}{Sheng Zhang},
  \bibinfo{person}{Lianwen Jin}, \bibinfo{person}{Lele Xie},
  \bibinfo{person}{Yaqiang Wu}, {and} \bibinfo{person}{Zhepeng Wang}.}
  \bibinfo{year}{2019}\natexlab{c}.
\newblock \showarticletitle{Omnidirectional scene text detection with
  sequential-free box discretization}. In \bibinfo{booktitle}{\emph{IJCAI}}.
  \bibinfo{pages}{3052--3058}.
\newblock


\bibitem[Liu et~al\mbox{.}(2019b)]%
        {liu2019CSE}
\bibfield{author}{\bibinfo{person}{Zichuan Liu}, \bibinfo{person}{Guosheng
  Lin}, \bibinfo{person}{Sheng Yang}, \bibinfo{person}{Fayao Liu},
  \bibinfo{person}{Weisi Lin}, {and} \bibinfo{person}{Wang~Ling Goh}.}
  \bibinfo{year}{2019}\natexlab{b}.
\newblock \showarticletitle{Towards robust curve text detection with
  conditional spatial expansion}. In \bibinfo{booktitle}{\emph{CVPR}}.
  \bibinfo{pages}{7269--7278}.
\newblock


\bibitem[Long et~al\mbox{.}(2015)]%
        {long2015FCN}
\bibfield{author}{\bibinfo{person}{Jonathan Long}, \bibinfo{person}{Evan
  Shelhamer}, {and} \bibinfo{person}{Trevor Darrell}.}
  \bibinfo{year}{2015}\natexlab{}.
\newblock \showarticletitle{Fully convolutional networks for semantic
  segmentation}. In \bibinfo{booktitle}{\emph{CVPR}}.
  \bibinfo{pages}{3431--3440}.
\newblock


\bibitem[Long et~al\mbox{.}(2021)]%
        {long2021era}
\bibfield{author}{\bibinfo{person}{Shangbang Long}, \bibinfo{person}{Xin He},
  {and} \bibinfo{person}{Cong Yao}.} \bibinfo{year}{2021}\natexlab{}.
\newblock \showarticletitle{Scene Text Detection and Recognition: The Deep
  Learning Era}.
\newblock \bibinfo{journal}{\emph{IJCV}} \bibinfo{volume}{129},
  \bibinfo{number}{1} (\bibinfo{year}{2021}), \bibinfo{pages}{161--184}.
\newblock


\bibitem[Long et~al\mbox{.}(2022)]%
        {long2022hiertext}
\bibfield{author}{\bibinfo{person}{Shangbang Long}, \bibinfo{person}{Siyang
  Qin}, \bibinfo{person}{Dmitry Panteleev}, \bibinfo{person}{Alessandro
  Bissacco}, \bibinfo{person}{Yasuhisa Fujii}, {and} \bibinfo{person}{Michalis
  Raptis}.} \bibinfo{year}{2022}\natexlab{}.
\newblock \showarticletitle{Towards end-to-end unified scene text detection and
  layout analysis}. In \bibinfo{booktitle}{\emph{CVPR}}.
  \bibinfo{pages}{1049--1059}.
\newblock


\bibitem[Long et~al\mbox{.}(2018)]%
        {long2018textsnake}
\bibfield{author}{\bibinfo{person}{Shangbang Long}, \bibinfo{person}{Jiaqiang
  Ruan}, \bibinfo{person}{Wenjie Zhang}, \bibinfo{person}{Xin He},
  \bibinfo{person}{Wenhao Wu}, {and} \bibinfo{person}{Cong Yao}.}
  \bibinfo{year}{2018}\natexlab{}.
\newblock \showarticletitle{Textsnake: A flexible representation for detecting
  text of arbitrary shapes}. In \bibinfo{booktitle}{\emph{ECCV}}.
  \bibinfo{pages}{20--36}.
\newblock


\bibitem[Lyu et~al\mbox{.}(2018)]%
        {lyu2018corner}
\bibfield{author}{\bibinfo{person}{Pengyuan Lyu}, \bibinfo{person}{Cong Yao},
  \bibinfo{person}{Wenhao Wu}, \bibinfo{person}{Shuicheng Yan}, {and}
  \bibinfo{person}{Xiang Bai}.} \bibinfo{year}{2018}\natexlab{}.
\newblock \showarticletitle{Multi-oriented scene text detection via corner
  localization and region segmentation}. In \bibinfo{booktitle}{\emph{CVPR}}.
  \bibinfo{pages}{7553--7563}.
\newblock


\bibitem[Ma et~al\mbox{.}(2021)]%
        {ma2021relatext}
\bibfield{author}{\bibinfo{person}{Chixiang Ma}, \bibinfo{person}{Lei Sun},
  \bibinfo{person}{Zhuoyao Zhong}, {and} \bibinfo{person}{Qiang Huo}.}
  \bibinfo{year}{2021}\natexlab{}.
\newblock \showarticletitle{ReLaText: Exploiting visual relationships for
  arbitrary-shaped scene text detection with graph convolutional networks}.
\newblock \bibinfo{journal}{\emph{PR}} (\bibinfo{year}{2021}),
  \bibinfo{pages}{337--345}.
\newblock


\bibitem[Ma et~al\mbox{.}(2018)]%
        {ma2018RRPN}
\bibfield{author}{\bibinfo{person}{Jianqi Ma}, \bibinfo{person}{Weiyuan Shao},
  \bibinfo{person}{Hao Ye}, \bibinfo{person}{Li Wang}, \bibinfo{person}{Hong
  Wang}, \bibinfo{person}{Yingbin Zheng}, {and} \bibinfo{person}{Xiangyang
  Xue}.} \bibinfo{year}{2018}\natexlab{}.
\newblock \showarticletitle{Arbitrary-oriented scene text detection via
  rotation proposals}.
\newblock \bibinfo{journal}{\emph{IEEE TMM}} (\bibinfo{year}{2018}),
  \bibinfo{pages}{3111--3122}.
\newblock


\bibitem[Milletari et~al\mbox{.}(2016)]%
        {milletari2016Vnet}
\bibfield{author}{\bibinfo{person}{Fausto Milletari}, \bibinfo{person}{Nassir
  Navab}, {and} \bibinfo{person}{Seyed-Ahmad Ahmadi}.}
  \bibinfo{year}{2016}\natexlab{}.
\newblock \showarticletitle{V-net: Fully convolutional neural networks for
  volumetric medical image segmentation}. In \bibinfo{booktitle}{\emph{3DV}}.
  \bibinfo{pages}{565--571}.
\newblock


\bibitem[Oord et~al\mbox{.}(2018)]%
        {oord2018cpc}
\bibfield{author}{\bibinfo{person}{Aaron van~den Oord}, \bibinfo{person}{Yazhe
  Li}, {and} \bibinfo{person}{Oriol Vinyals}.} \bibinfo{year}{2018}\natexlab{}.
\newblock \showarticletitle{Representation learning with contrastive predictive
  coding}.
\newblock \bibinfo{journal}{\emph{arXiv preprint arXiv:1807.03748}}
  (\bibinfo{year}{2018}).
\newblock


\bibitem[Qiao et~al\mbox{.}(2020a)]%
        {qiao2020gcan}
\bibfield{author}{\bibinfo{person}{Zhi Qiao}, \bibinfo{person}{Xugong Qin},
  \bibinfo{person}{Yu Zhou}, \bibinfo{person}{Fei Yang}, {and}
  \bibinfo{person}{Weiping Wang}.} \bibinfo{year}{2020}\natexlab{a}.
\newblock \showarticletitle{Gaussian Constrained Attention Network for Scene
  Text Recognition}. In \bibinfo{booktitle}{\emph{{ICPR}}}.
  \bibinfo{pages}{3328--3335}.
\newblock


\bibitem[Qiao et~al\mbox{.}(2021)]%
        {qiao2021pimnet}
\bibfield{author}{\bibinfo{person}{Zhi Qiao}, \bibinfo{person}{Yu Zhou},
  \bibinfo{person}{Jin Wei}, \bibinfo{person}{Wei Wang}, \bibinfo{person}{Yuan
  Zhang}, \bibinfo{person}{Ning Jiang}, \bibinfo{person}{Hongbin Wang}, {and}
  \bibinfo{person}{Weiping Wang}.} \bibinfo{year}{2021}\natexlab{}.
\newblock \showarticletitle{PIMNet: {A} Parallel, Iterative and Mimicking
  Network for Scene Text Recognition}. In \bibinfo{booktitle}{\emph{ACM MM}}.
  \bibinfo{pages}{2046--2055}.
\newblock


\bibitem[Qiao et~al\mbox{.}(2020b)]%
        {qiao2020seed}
\bibfield{author}{\bibinfo{person}{Zhi Qiao}, \bibinfo{person}{Yu Zhou},
  \bibinfo{person}{Dongbao Yang}, \bibinfo{person}{Yucan Zhou}, {and}
  \bibinfo{person}{Weiping Wang}.} \bibinfo{year}{2020}\natexlab{b}.
\newblock \showarticletitle{{SEED:} Semantics Enhanced Encoder-Decoder
  Framework for Scene Text Recognition}. In \bibinfo{booktitle}{\emph{CVPR}}.
  \bibinfo{pages}{13525--13534}.
\newblock


\bibitem[Qin et~al\mbox{.}(2021b)]%
        {qin2021mayor}
\bibfield{author}{\bibinfo{person}{Xugong Qin}, \bibinfo{person}{Yu Zhou},
  \bibinfo{person}{Youhui Guo}, \bibinfo{person}{Dayan Wu},
  \bibinfo{person}{Zhihong Tian}, \bibinfo{person}{Ning Jiang},
  \bibinfo{person}{Hongbin Wang}, {and} \bibinfo{person}{Weiping Wang}.}
  \bibinfo{year}{2021}\natexlab{b}.
\newblock \showarticletitle{Mask is all you need: Rethinking mask R-CNN for
  dense and arbitrary-shaped scene text detection}. In
  \bibinfo{booktitle}{\emph{ACM MM}}. \bibinfo{pages}{414--423}.
\newblock


\bibitem[Qin et~al\mbox{.}(2021a)]%
        {qin2021fc2rn}
\bibfield{author}{\bibinfo{person}{Xugong Qin}, \bibinfo{person}{Yu Zhou},
  \bibinfo{person}{Youhui Guo}, \bibinfo{person}{Dayan Wu}, {and}
  \bibinfo{person}{Weiping Wang}.} \bibinfo{year}{2021}\natexlab{a}.
\newblock \showarticletitle{FC\({}^{\mbox{2}}\)RN: {A} Fully Convolutional
  Corner Refinement Network for Accurate Multi-Oriented Scene Text Detection}.
  In \bibinfo{booktitle}{\emph{{ICASSP}}}. \bibinfo{pages}{4350--4354}.
\newblock


\bibitem[Qin et~al\mbox{.}(2019)]%
        {qin2019wsctd}
\bibfield{author}{\bibinfo{person}{Xugong Qin}, \bibinfo{person}{Yu Zhou},
  \bibinfo{person}{Dongbao Yang}, {and} \bibinfo{person}{Weiping Wang}.}
  \bibinfo{year}{2019}\natexlab{}.
\newblock \showarticletitle{Curved Text Detection in Natural Scene Images with
  Semi- and Weakly-Supervised Learning}. In
  \bibinfo{booktitle}{\emph{{ICDAR}}}. \bibinfo{pages}{559--564}.
\newblock


\bibitem[Ren et~al\mbox{.}(2015)]%
        {ren2015fasterrcnn}
\bibfield{author}{\bibinfo{person}{Shaoqing Ren}, \bibinfo{person}{Kaiming He},
  \bibinfo{person}{Ross Girshick}, {and} \bibinfo{person}{Jian Sun}.}
  \bibinfo{year}{2015}\natexlab{}.
\newblock \showarticletitle{Faster R-CNN: Towards real-time object detection
  with region proposal networks}. In \bibinfo{booktitle}{\emph{NeurIPS}}.
  \bibinfo{pages}{91--99}.
\newblock


\bibitem[Shen et~al\mbox{.}(2023)]%
        {shen2023divide}
\bibfield{author}{\bibinfo{person}{Huawen Shen}, \bibinfo{person}{Xiang Gao},
  \bibinfo{person}{Jin Wei}, \bibinfo{person}{Liang Qiao}, \bibinfo{person}{Yu
  Zhou}, \bibinfo{person}{Qiang Li}, {and} \bibinfo{person}{Zhanzhan Cheng}.}
  \bibinfo{year}{2023}\natexlab{}.
\newblock \showarticletitle{Divide Rows and Conquer Cells: Towards Structure
  Recognition for Large Tables}. In \bibinfo{booktitle}{\emph{IJCAI}}.
\newblock


\bibitem[Sheng et~al\mbox{.}(2021)]%
        {sheng2021centripetaltext}
\bibfield{author}{\bibinfo{person}{Tao Sheng}, \bibinfo{person}{Jie Chen},
  {and} \bibinfo{person}{Zhouhui Lian}.} \bibinfo{year}{2021}\natexlab{}.
\newblock \showarticletitle{Centripetaltext: An efficient text instance
  representation for scene text detection}. In
  \bibinfo{booktitle}{\emph{NeurIPS}}. \bibinfo{pages}{335--346}.
\newblock


\bibitem[Shi et~al\mbox{.}(2017a)]%
        {shi2017SegLink}
\bibfield{author}{\bibinfo{person}{Baoguang Shi}, \bibinfo{person}{Xiang Bai},
  {and} \bibinfo{person}{Serge Belongie}.} \bibinfo{year}{2017}\natexlab{a}.
\newblock \showarticletitle{Detecting oriented text in natural images by
  linking segments}. In \bibinfo{booktitle}{\emph{CVPR}}.
  \bibinfo{pages}{2550--2558}.
\newblock


\bibitem[Shi et~al\mbox{.}(2017b)]%
        {shi2016crnn}
\bibfield{author}{\bibinfo{person}{Baoguang Shi}, \bibinfo{person}{Xiang Bai},
  {and} \bibinfo{person}{Cong Yao}.} \bibinfo{year}{2017}\natexlab{b}.
\newblock \showarticletitle{An end-to-end trainable neural network for
  image-based sequence recognition and its application to scene text
  recognition}.
\newblock \bibinfo{journal}{\emph{IEEE TPAMI}} \bibinfo{volume}{39},
  \bibinfo{number}{11} (\bibinfo{year}{2017}), \bibinfo{pages}{2298--2304}.
\newblock


\bibitem[Shi et~al\mbox{.}(2019)]%
        {shi2018aster}
\bibfield{author}{\bibinfo{person}{Baoguang Shi}, \bibinfo{person}{Mingkun
  Yang}, \bibinfo{person}{Xinggang Wang}, \bibinfo{person}{Pengyuan Lyu},
  \bibinfo{person}{Cong Yao}, {and} \bibinfo{person}{Xiang Bai}.}
  \bibinfo{year}{2019}\natexlab{}.
\newblock \showarticletitle{Aster: An attentional scene text recognizer with
  flexible rectification}.
\newblock \bibinfo{journal}{\emph{IEEE TPAMI}} \bibinfo{volume}{41},
  \bibinfo{number}{9} (\bibinfo{year}{2019}), \bibinfo{pages}{2035--2048}.
\newblock


\bibitem[Song et~al\mbox{.}(2022)]%
        {song2022VLPT}
\bibfield{author}{\bibinfo{person}{Sibo Song}, \bibinfo{person}{Jianqiang Wan},
  \bibinfo{person}{Zhibo Yang}, \bibinfo{person}{Jun Tang},
  \bibinfo{person}{Wenqing Cheng}, \bibinfo{person}{Xiang Bai}, {and}
  \bibinfo{person}{Cong Yao}.} \bibinfo{year}{2022}\natexlab{}.
\newblock \showarticletitle{Vision-language pre-training for boosting scene
  text detectors}. In \bibinfo{booktitle}{\emph{CVPR}}.
  \bibinfo{pages}{15681--15691}.
\newblock


\bibitem[Tang et~al\mbox{.}(2019)]%
        {tang2019seglink++}
\bibfield{author}{\bibinfo{person}{Jun Tang}, \bibinfo{person}{Zhibo Yang},
  \bibinfo{person}{Yongpan Wang}, \bibinfo{person}{Qi Zheng},
  \bibinfo{person}{Yongchao Xu}, {and} \bibinfo{person}{Xiang Bai}.}
  \bibinfo{year}{2019}\natexlab{}.
\newblock \showarticletitle{SegLink++: Detecting dense and arbitrary-shaped
  scene text by instance-aware component grouping}.
\newblock \bibinfo{journal}{\emph{PR}} (\bibinfo{year}{2019}),
  \bibinfo{pages}{106954}.
\newblock


\bibitem[Tang et~al\mbox{.}(2022)]%
        {tang2022FSG}
\bibfield{author}{\bibinfo{person}{Jingqun Tang}, \bibinfo{person}{Wenqing
  Zhang}, \bibinfo{person}{Hongye Liu}, \bibinfo{person}{Mingkun Yang},
  \bibinfo{person}{Bo Jiang}, \bibinfo{person}{Guanglong Hu}, {and}
  \bibinfo{person}{Xiang Bai}.} \bibinfo{year}{2022}\natexlab{}.
\newblock \showarticletitle{Few Could Be Better Than All: Feature Sampling and
  Grouping for Scene Text Detection}. In \bibinfo{booktitle}{\emph{CVPR}}.
  \bibinfo{publisher}{{IEEE}}.
\newblock


\bibitem[Tian et~al\mbox{.}(2016)]%
        {tian2016CTPN}
\bibfield{author}{\bibinfo{person}{Zhi Tian}, \bibinfo{person}{Weilin Huang},
  \bibinfo{person}{Tong He}, \bibinfo{person}{Pan He}, {and}
  \bibinfo{person}{Yu Qiao}.} \bibinfo{year}{2016}\natexlab{}.
\newblock \showarticletitle{Detecting text in natural image with connectionist
  text proposal network}. In \bibinfo{booktitle}{\emph{ECCV}}.
  \bibinfo{pages}{56--72}.
\newblock


\bibitem[Tian et~al\mbox{.}(2019)]%
        {tian2019SAE}
\bibfield{author}{\bibinfo{person}{Zhuotao Tian}, \bibinfo{person}{Michelle
  Shu}, \bibinfo{person}{Pengyuan Lyu}, \bibinfo{person}{Ruiyu Li},
  \bibinfo{person}{Chao Zhou}, \bibinfo{person}{Xiaoyong Shen}, {and}
  \bibinfo{person}{Jiaya Jia}.} \bibinfo{year}{2019}\natexlab{}.
\newblock \showarticletitle{Learning shape-aware embedding for scene text
  detection}. In \bibinfo{booktitle}{\emph{CVPR}}. \bibinfo{pages}{4234--4243}.
\newblock


\bibitem[Vatti(1992)]%
        {vatti1992vatti}
\bibfield{author}{\bibinfo{person}{Bala~R Vatti}.}
  \bibinfo{year}{1992}\natexlab{}.
\newblock \showarticletitle{A generic solution to polygon clipping}.
\newblock \bibinfo{journal}{\emph{Commun. ACM}} \bibinfo{volume}{35},
  \bibinfo{number}{7} (\bibinfo{year}{1992}), \bibinfo{pages}{56--63}.
\newblock


\bibitem[Wan et~al\mbox{.}(2021)]%
        {wan2021STKD}
\bibfield{author}{\bibinfo{person}{Qi Wan}, \bibinfo{person}{Haoqin Ji}, {and}
  \bibinfo{person}{Linlin Shen}.} \bibinfo{year}{2021}\natexlab{}.
\newblock \showarticletitle{Self-attention based text knowledge mining for text
  detection}. In \bibinfo{booktitle}{\emph{CVPR}}. \bibinfo{pages}{5983--5992}.
\newblock


\bibitem[Wang et~al\mbox{.}(2020a)]%
        {Wang2020textray}
\bibfield{author}{\bibinfo{person}{Fangfang Wang}, \bibinfo{person}{Yifeng
  Chen}, \bibinfo{person}{Fei Wu}, {and} \bibinfo{person}{Xi Li}.}
  \bibinfo{year}{2020}\natexlab{a}.
\newblock \showarticletitle{TextRay: Contour-based geometric modeling for
  arbitrary-shaped scene text detection}. In \bibinfo{booktitle}{\emph{ACM
  MM}}. \bibinfo{pages}{111--119}.
\newblock


\bibitem[Wang et~al\mbox{.}(2018)]%
        {wang2018ITN}
\bibfield{author}{\bibinfo{person}{Fangfang Wang}, \bibinfo{person}{Liming
  Zhao}, \bibinfo{person}{Xi Li}, \bibinfo{person}{Xinchao Wang}, {and}
  \bibinfo{person}{Dacheng Tao}.} \bibinfo{year}{2018}\natexlab{}.
\newblock \showarticletitle{Geometry-aware scene text detection with instance
  transformation network}. In \bibinfo{booktitle}{\emph{CVPR}}.
  \bibinfo{pages}{1381--1389}.
\newblock


\bibitem[Wang et~al\mbox{.}(2022a)]%
        {wang2022kmfgr}
\bibfield{author}{\bibinfo{person}{Hao Wang}, \bibinfo{person}{Junchao Liao},
  \bibinfo{person}{Tianheng Cheng}, \bibinfo{person}{Zewen Gao},
  \bibinfo{person}{Hao Liu}, \bibinfo{person}{Bo Ren}, \bibinfo{person}{Xiang
  Bai}, {and} \bibinfo{person}{Wenyu Liu}.} \bibinfo{year}{2022}\natexlab{a}.
\newblock \showarticletitle{Knowledge mining with scene text for fine-grained
  recognition}. In \bibinfo{booktitle}{\emph{CVPR}}.
  \bibinfo{pages}{4624--4633}.
\newblock


\bibitem[Wang et~al\mbox{.}(2019d)]%
        {wang2019SAST}
\bibfield{author}{\bibinfo{person}{Pengfei Wang}, \bibinfo{person}{Chengquan
  Zhang}, \bibinfo{person}{Fei Qi}, \bibinfo{person}{Zuming Huang},
  \bibinfo{person}{Mengyi En}, \bibinfo{person}{Junyu Han},
  \bibinfo{person}{Jingtuo Liu}, \bibinfo{person}{Errui Ding}, {and}
  \bibinfo{person}{Guangming Shi}.} \bibinfo{year}{2019}\natexlab{d}.
\newblock \showarticletitle{A single-shot arbitrarily-shaped text detector
  based on context attended multi-task learning}. In
  \bibinfo{booktitle}{\emph{ACM MM}}. \bibinfo{pages}{1277--1285}.
\newblock


\bibitem[Wang et~al\mbox{.}(2019b)]%
        {wang2019PSENet}
\bibfield{author}{\bibinfo{person}{Wenhai Wang}, \bibinfo{person}{Enze Xie},
  \bibinfo{person}{Xiang Li}, \bibinfo{person}{Wenbo Hou},
  \bibinfo{person}{Tong Lu}, \bibinfo{person}{Gang Yu}, {and}
  \bibinfo{person}{Shuai Shao}.} \bibinfo{year}{2019}\natexlab{b}.
\newblock \showarticletitle{Shape robust text detection with progressive scale
  expansion network}. In \bibinfo{booktitle}{\emph{CVPR}}.
  \bibinfo{pages}{9336--9345}.
\newblock


\bibitem[Wang et~al\mbox{.}(2019c)]%
        {wang2019PAN}
\bibfield{author}{\bibinfo{person}{Wenhai Wang}, \bibinfo{person}{Enze Xie},
  \bibinfo{person}{Xiaoge Song}, \bibinfo{person}{Yuhang Zang},
  \bibinfo{person}{Wenjia Wang}, \bibinfo{person}{Tong Lu},
  \bibinfo{person}{Gang Yu}, {and} \bibinfo{person}{Chunhua Shen}.}
  \bibinfo{year}{2019}\natexlab{c}.
\newblock \showarticletitle{Efficient and accurate arbitrary-shaped text
  detection with pixel aggregation network}. In
  \bibinfo{booktitle}{\emph{ICCV}}. \bibinfo{pages}{8440--8449}.
\newblock


\bibitem[Wang et~al\mbox{.}(2022c)]%
        {wang2022tpsnet}
\bibfield{author}{\bibinfo{person}{Wei Wang}, \bibinfo{person}{Yu Zhou},
  \bibinfo{person}{Jiahao Lv}, \bibinfo{person}{Dayan Wu},
  \bibinfo{person}{Guoqing Zhao}, \bibinfo{person}{Ning Jiang}, {and}
  \bibinfo{person}{Weipinng Wang}.} \bibinfo{year}{2022}\natexlab{c}.
\newblock \showarticletitle{Tpsnet: Reverse thinking of thin plate splines for
  arbitrary shape scene text representation}. In \bibinfo{booktitle}{\emph{ACM
  MM}}. \bibinfo{pages}{5014--5025}.
\newblock


\bibitem[Wang et~al\mbox{.}(2019a)]%
        {wang2019ATRR}
\bibfield{author}{\bibinfo{person}{Xiaobing Wang}, \bibinfo{person}{Yingying
  Jiang}, \bibinfo{person}{Zhenbo Luo}, \bibinfo{person}{Cheng-Lin Liu},
  \bibinfo{person}{Hyunsoo Choi}, {and} \bibinfo{person}{Sungjin Kim}.}
  \bibinfo{year}{2019}\natexlab{a}.
\newblock \showarticletitle{Arbitrary shape scene text detection with adaptive
  text region representation}. In \bibinfo{booktitle}{\emph{CVPR}}.
  \bibinfo{pages}{6449--6458}.
\newblock


\bibitem[Wang et~al\mbox{.}(2020c)]%
        {wang2020solov2}
\bibfield{author}{\bibinfo{person}{Xinlong Wang}, \bibinfo{person}{Rufeng
  Zhang}, \bibinfo{person}{Tao Kong}, \bibinfo{person}{Lei Li}, {and}
  \bibinfo{person}{Chunhua Shen}.} \bibinfo{year}{2020}\natexlab{c}.
\newblock \showarticletitle{Solov2: Dynamic and fast instance segmentation}.
\newblock \bibinfo{journal}{\emph{NeurIPS}} (\bibinfo{year}{2020}),
  \bibinfo{pages}{17721--17732}.
\newblock


\bibitem[Wang et~al\mbox{.}(2022b)]%
        {wang2022TSTD}
\bibfield{author}{\bibinfo{person}{Yuxin Wang}, \bibinfo{person}{Hongtao Xie},
  \bibinfo{person}{Mengting Xing}, \bibinfo{person}{Jing Wang},
  \bibinfo{person}{Shenggao Zhu}, {and} \bibinfo{person}{Yongdong Zhang}.}
  \bibinfo{year}{2022}\natexlab{b}.
\newblock \showarticletitle{Detecting Tampered Scene Text in the Wild}. In
  \bibinfo{booktitle}{\emph{ECCV}}. \bibinfo{pages}{215--232}.
\newblock


\bibitem[Wang et~al\mbox{.}(2020b)]%
        {Wang2020contournet}
\bibfield{author}{\bibinfo{person}{Yuxin Wang}, \bibinfo{person}{Hongtao Xie},
  \bibinfo{person}{Zheng-Jun Zha}, \bibinfo{person}{Mengting Xing},
  \bibinfo{person}{Zilong Fu}, {and} \bibinfo{person}{Yongdong Zhang}.}
  \bibinfo{year}{2020}\natexlab{b}.
\newblock \showarticletitle{ContourNet: Taking a further step toward accurate
  arbitrary-shaped scene text detection}. In \bibinfo{booktitle}{\emph{CVPR}}.
  \bibinfo{pages}{11750--11759}.
\newblock


\bibitem[Wei et~al\mbox{.}(2022)]%
        {wj2022textblock}
\bibfield{author}{\bibinfo{person}{Jin Wei}, \bibinfo{person}{Yuan Zhang},
  \bibinfo{person}{Yu Zhou}, \bibinfo{person}{Gangyan Zeng},
  \bibinfo{person}{Zhi Qiao}, \bibinfo{person}{Youhui Guo},
  \bibinfo{person}{Haiying Wu}, \bibinfo{person}{Hongbin Wang}, {and}
  \bibinfo{person}{Weiping Wang}.} \bibinfo{year}{2022}\natexlab{}.
\newblock \showarticletitle{TextBlock: Towards Scene Text Spotting without
  Fine-Grained Detection}. In \bibinfo{booktitle}{\emph{ACM MM}}.
  \bibinfo{pages}{5892–5902}.
\newblock


\bibitem[Wu and Natarajan(2017)]%
        {wu2017self}
\bibfield{author}{\bibinfo{person}{Yue Wu} {and} \bibinfo{person}{Prem
  Natarajan}.} \bibinfo{year}{2017}\natexlab{}.
\newblock \showarticletitle{Self-organized text detection with minimal
  post-processing via border learning}. In \bibinfo{booktitle}{\emph{ICCV}}.
  \bibinfo{pages}{5000--5009}.
\newblock


\bibitem[Xiao et~al\mbox{.}(2020)]%
        {xiao2020SD}
\bibfield{author}{\bibinfo{person}{Shanyu Xiao}, \bibinfo{person}{Liangrui
  Peng}, \bibinfo{person}{Yan Ruijie}, \bibinfo{person}{An Keyu},
  \bibinfo{person}{Yao Gang}, {and} \bibinfo{person}{Min Jaesik}.}
  \bibinfo{year}{2020}\natexlab{}.
\newblock \showarticletitle{Sequential deformation for accurate scene text
  detection}. In \bibinfo{booktitle}{\emph{ECCV}}. \bibinfo{pages}{108--124}.
\newblock


\bibitem[Xie et~al\mbox{.}(2019)]%
        {xie2019SPCNet}
\bibfield{author}{\bibinfo{person}{Enze Xie}, \bibinfo{person}{Yuhang Zang},
  \bibinfo{person}{Shuai Shao}, \bibinfo{person}{Gang Yu},
  \bibinfo{person}{Cong Yao}, {and} \bibinfo{person}{Guangyao Li}.}
  \bibinfo{year}{2019}\natexlab{}.
\newblock \showarticletitle{Scene text detection with supervised pyramid
  context network}. In \bibinfo{booktitle}{\emph{AAAI}}.
  \bibinfo{pages}{9038--9045}.
\newblock


\bibitem[Xing et~al\mbox{.}(2019)]%
        {xing2019CCN}
\bibfield{author}{\bibinfo{person}{Linjie Xing}, \bibinfo{person}{Zhi Tian},
  \bibinfo{person}{Weilin Huang}, {and} \bibinfo{person}{Matthew~R Scott}.}
  \bibinfo{year}{2019}\natexlab{}.
\newblock \showarticletitle{Convolutional character networks}. In
  \bibinfo{booktitle}{\emph{ICCV}}. \bibinfo{pages}{9126--9136}.
\newblock


\bibitem[Xu et~al\mbox{.}(2019)]%
        {xu2019textfield}
\bibfield{author}{\bibinfo{person}{Yongchao Xu}, \bibinfo{person}{Yukang Wang},
  \bibinfo{person}{Wei Zhou}, \bibinfo{person}{Yongpan Wang},
  \bibinfo{person}{Zhibo Yang}, {and} \bibinfo{person}{Xiang Bai}.}
  \bibinfo{year}{2019}\natexlab{}.
\newblock \showarticletitle{TextField: Learning a deep direction field for
  irregular scene text detection}.
\newblock \bibinfo{journal}{\emph{IEEE TIP}} (\bibinfo{year}{2019}),
  \bibinfo{pages}{5566--5579}.
\newblock


\bibitem[Xue et~al\mbox{.}(2018)]%
        {xue2018border}
\bibfield{author}{\bibinfo{person}{Chuhui Xue}, \bibinfo{person}{Shijian Lu},
  {and} \bibinfo{person}{Fangneng Zhan}.} \bibinfo{year}{2018}\natexlab{}.
\newblock \showarticletitle{Accurate scene text detection through border
  semantics awareness and bootstrapping}. In \bibinfo{booktitle}{\emph{ECCV}}.
  \bibinfo{pages}{355--372}.
\newblock


\bibitem[Xue et~al\mbox{.}(2019)]%
        {xue2019msr}
\bibfield{author}{\bibinfo{person}{Chuhui Xue}, \bibinfo{person}{Shijian Lu},
  {and} \bibinfo{person}{Wei Zhang}.} \bibinfo{year}{2019}\natexlab{}.
\newblock \showarticletitle{MSR: multi-scale shape regression for scene text
  detection}. In \bibinfo{booktitle}{\emph{IJCAI}}. \bibinfo{pages}{989--995}.
\newblock


\bibitem[Xue et~al\mbox{.}(2022)]%
        {xue2022oCLIP}
\bibfield{author}{\bibinfo{person}{Chuhui Xue}, \bibinfo{person}{Wenqing
  Zhang}, \bibinfo{person}{Yu Hao}, \bibinfo{person}{Shijian Lu},
  \bibinfo{person}{Philip~HS Torr}, {and} \bibinfo{person}{Song Bai}.}
  \bibinfo{year}{2022}\natexlab{}.
\newblock \showarticletitle{Language matters: A weakly supervised
  vision-language pre-training approach for scene text detection and spotting}.
  In \bibinfo{booktitle}{\emph{ECCV}}. \bibinfo{pages}{284--302}.
\newblock


\bibitem[Yang et~al\mbox{.}(2018)]%
        {yang2018inceptext}
\bibfield{author}{\bibinfo{person}{Qiangpeng Yang}, \bibinfo{person}{Mengli
  Cheng}, \bibinfo{person}{Wenmeng Zhou}, \bibinfo{person}{Yan Chen},
  \bibinfo{person}{Minghui Qiu}, {and} \bibinfo{person}{Wei Lin}.}
  \bibinfo{year}{2018}\natexlab{}.
\newblock \showarticletitle{Inceptext: A new inception-text module with
  deformable PSROI pooling for multi-oriented scene text detection}. In
  \bibinfo{booktitle}{\emph{IJCAI}}. \bibinfo{pages}{1071--1077}.
\newblock


\bibitem[Yao et~al\mbox{.}(2014a)]%
        {yao2014unified}
\bibfield{author}{\bibinfo{person}{Cong Yao}, \bibinfo{person}{Xiang Bai},
  {and} \bibinfo{person}{Wenyu Liu}.} \bibinfo{year}{2014}\natexlab{a}.
\newblock \showarticletitle{A unified framework for multioriented text
  detection and recognition}.
\newblock \bibinfo{journal}{\emph{IEEE TIP}} (\bibinfo{year}{2014}),
  \bibinfo{pages}{4737--4749}.
\newblock


\bibitem[Yao et~al\mbox{.}(2012)]%
        {yao2012detecting}
\bibfield{author}{\bibinfo{person}{Cong Yao}, \bibinfo{person}{Xiang Bai},
  \bibinfo{person}{Wenyu Liu}, \bibinfo{person}{Yi Ma}, {and}
  \bibinfo{person}{Zhuowen Tu}.} \bibinfo{year}{2012}\natexlab{}.
\newblock \showarticletitle{Detecting texts of arbitrary orientations in
  natural images}. In \bibinfo{booktitle}{\emph{CVPR}}.
  \bibinfo{pages}{1083--1090}.
\newblock


\bibitem[Yao et~al\mbox{.}(2016)]%
        {yao2016holistic}
\bibfield{author}{\bibinfo{person}{Cong Yao}, \bibinfo{person}{Xiang Bai},
  \bibinfo{person}{Nong Sang}, \bibinfo{person}{Xinyu Zhou},
  \bibinfo{person}{Shuchang Zhou}, {and} \bibinfo{person}{Zhimin Cao}.}
  \bibinfo{year}{2016}\natexlab{}.
\newblock \showarticletitle{Scene text detection via holistic, multi-channel
  prediction}.
\newblock \bibinfo{journal}{\emph{arXiv preprint arXiv:1606.09002}}
  (\bibinfo{year}{2016}).
\newblock


\bibitem[Yao et~al\mbox{.}(2014b)]%
        {yao2014strokelets}
\bibfield{author}{\bibinfo{person}{Cong Yao}, \bibinfo{person}{Xiang Bai},
  \bibinfo{person}{Baoguang Shi}, {and} \bibinfo{person}{Wenyu Liu}.}
  \bibinfo{year}{2014}\natexlab{b}.
\newblock \showarticletitle{Strokelets: A learned multi-scale representation
  for scene text recognition}. In \bibinfo{booktitle}{\emph{CVPR}}.
  \bibinfo{pages}{4042--4049}.
\newblock


\bibitem[Ye et~al\mbox{.}(2020)]%
        {ye2020textfusenet}
\bibfield{author}{\bibinfo{person}{Jian Ye}, \bibinfo{person}{Zhe Chen},
  \bibinfo{person}{Juhua Liu}, {and} \bibinfo{person}{Bo Du}.}
  \bibinfo{year}{2020}\natexlab{}.
\newblock \showarticletitle{TextFuseNet: Scene text detection with richer fused
  features}. In \bibinfo{booktitle}{\emph{IJCAI}}. \bibinfo{pages}{516--522}.
\newblock


\bibitem[Ye and Doermann(2015)]%
        {ye2015survey}
\bibfield{author}{\bibinfo{person}{Qixiang Ye} {and} \bibinfo{person}{David~S.
  Doermann}.} \bibinfo{year}{2015}\natexlab{}.
\newblock \showarticletitle{Text Detection and Recognition in Imagery: {A}
  Survey}.
\newblock \bibinfo{journal}{\emph{IEEE TPAMI}} \bibinfo{volume}{37},
  \bibinfo{number}{7} (\bibinfo{year}{2015}), \bibinfo{pages}{1480--1500}.
\newblock


\bibitem[Yu et~al\mbox{.}(2023)]%
        {yu2023TCM}
\bibfield{author}{\bibinfo{person}{Wenwen Yu}, \bibinfo{person}{Yuliang Liu},
  \bibinfo{person}{Wei Hua}, \bibinfo{person}{Deqiang Jiang},
  \bibinfo{person}{Bo Ren}, {and} \bibinfo{person}{Xiang Bai}.}
  \bibinfo{year}{2023}\natexlab{}.
\newblock \showarticletitle{Turning a CLIP Model into a Scene Text Detector}.
\newblock \bibinfo{journal}{\emph{CVPR}}, \bibinfo{pages}{6978--6988}.
\newblock


\bibitem[Zeng et~al\mbox{.}(2021)]%
        {zgy2021mm}
\bibfield{author}{\bibinfo{person}{Gangyan Zeng}, \bibinfo{person}{Yuan Zhang},
  \bibinfo{person}{Yu Zhou}, {and} \bibinfo{person}{Xiaomeng Yang}.}
  \bibinfo{year}{2021}\natexlab{}.
\newblock \showarticletitle{Beyond {OCR} + {VQA:} Involving {OCR} into the Flow
  for Robust and Accurate TextVQA}. In \bibinfo{booktitle}{\emph{ACM MM}}.
  \bibinfo{pages}{376--385}.
\newblock


\bibitem[Zeng et~al\mbox{.}(2023)]%
        {zgy2022pr}
\bibfield{author}{\bibinfo{person}{Gangyan Zeng}, \bibinfo{person}{Yuan Zhang},
  \bibinfo{person}{Yu Zhou}, \bibinfo{person}{Xiaomeng Yang},
  \bibinfo{person}{Ning Jiang}, \bibinfo{person}{Guoqing Zhao},
  \bibinfo{person}{Weiping Wang}, {and} \bibinfo{person}{Xu-Cheng Yin}.}
  \bibinfo{year}{2023}\natexlab{}.
\newblock \showarticletitle{Beyond OCR+ VQA: Towards end-to-end reading and
  reasoning for robust and accurate textvqa}.
\newblock \bibinfo{journal}{\emph{PR}}  \bibinfo{volume}{138}
  (\bibinfo{year}{2023}), \bibinfo{pages}{109337}.
\newblock


\bibitem[Zhang et~al\mbox{.}(2019)]%
        {zhang2019lomo}
\bibfield{author}{\bibinfo{person}{Chengquan Zhang}, \bibinfo{person}{Borong
  Liang}, \bibinfo{person}{Zuming Huang}, \bibinfo{person}{Mengyi En},
  \bibinfo{person}{Junyu Han}, \bibinfo{person}{Errui Ding}, {and}
  \bibinfo{person}{Xinghao Ding}.} \bibinfo{year}{2019}\natexlab{}.
\newblock \showarticletitle{Look more than once: An accurate detector for text
  of arbitrary shapes}. In \bibinfo{booktitle}{\emph{CVPR}}.
  \bibinfo{pages}{10552--10561}.
\newblock


\bibitem[Zhang et~al\mbox{.}(2020a)]%
        {Zhang2020ATSS}
\bibfield{author}{\bibinfo{person}{Shifeng Zhang}, \bibinfo{person}{Cheng Chi},
  \bibinfo{person}{Yongqiang Yao}, \bibinfo{person}{Zhen Lei}, {and}
  \bibinfo{person}{Stan~Z. Li}.} \bibinfo{year}{2020}\natexlab{a}.
\newblock \showarticletitle{Bridging the gap between anchor-based and
  anchor-free detection via adaptive training sample selection}. In
  \bibinfo{booktitle}{\emph{CVPR}}. \bibinfo{pages}{9756--9765}.
\newblock


\bibitem[Zhang et~al\mbox{.}(2020b)]%
        {zhang2020DRRG}
\bibfield{author}{\bibinfo{person}{Shi-Xue Zhang}, \bibinfo{person}{Xiaobin
  Zhu}, \bibinfo{person}{Jie-Bo Hou}, \bibinfo{person}{Chang Liu},
  \bibinfo{person}{Chun Yang}, \bibinfo{person}{Hongfa Wang}, {and}
  \bibinfo{person}{Xu-Cheng Yin}.} \bibinfo{year}{2020}\natexlab{b}.
\newblock \showarticletitle{Deep relational reasoning graph network for
  arbitrary shape text detection}. In \bibinfo{booktitle}{\emph{CVPR}}.
  \bibinfo{pages}{9696--9705}.
\newblock


\bibitem[Zhang et~al\mbox{.}(2021)]%
        {zhang2021textbpn}
\bibfield{author}{\bibinfo{person}{Shi-Xue Zhang}, \bibinfo{person}{Xiaobin
  Zhu}, \bibinfo{person}{Chun Yang}, \bibinfo{person}{Hongfa Wang}, {and}
  \bibinfo{person}{Xu-Cheng Yin}.} \bibinfo{year}{2021}\natexlab{}.
\newblock \showarticletitle{Adaptive Boundary Proposal Network for Arbitrary
  Shape Text Detection}. In \bibinfo{booktitle}{\emph{ICCV}}.
  \bibinfo{pages}{1305--1314}.
\newblock


\bibitem[Zhang et~al\mbox{.}(2016)]%
        {Zhang2016FCNTD}
\bibfield{author}{\bibinfo{person}{Zheng Zhang}, \bibinfo{person}{Chengquan
  Zhang}, \bibinfo{person}{Wei Shen}, \bibinfo{person}{Cong Yao},
  \bibinfo{person}{Wenyu Liu}, {and} \bibinfo{person}{Xiang Bai}.}
  \bibinfo{year}{2016}\natexlab{}.
\newblock \showarticletitle{Multi-oriented text detection with fully
  convolutional networks}. In \bibinfo{booktitle}{\emph{CVPR}}.
  \bibinfo{pages}{4159--4167}.
\newblock


\bibitem[Zhou et~al\mbox{.}(2017)]%
        {zhou2017east}
\bibfield{author}{\bibinfo{person}{Xinyu Zhou}, \bibinfo{person}{Cong Yao},
  \bibinfo{person}{He Wen}, \bibinfo{person}{Yuzhi Wang},
  \bibinfo{person}{Shuchang Zhou}, \bibinfo{person}{Weiran He}, {and}
  \bibinfo{person}{Jiajun Liang}.} \bibinfo{year}{2017}\natexlab{}.
\newblock \showarticletitle{{EAST}: An efficient and accurate scene text
  detector}. In \bibinfo{booktitle}{\emph{CVPR}}. \bibinfo{pages}{2642--2651}.
\newblock


\bibitem[Zhou et~al\mbox{.}(2020)]%
        {zhou2020crnet}
\bibfield{author}{\bibinfo{person}{Yu Zhou}, \bibinfo{person}{Hongtao Xie},
  \bibinfo{person}{Shancheng Fang}, \bibinfo{person}{Yan Li}, {and}
  \bibinfo{person}{Yongdong Zhang}.} \bibinfo{year}{2020}\natexlab{}.
\newblock \showarticletitle{CRNet: A center-aware representation for detecting
  text of arbitrary shapes}. In \bibinfo{booktitle}{\emph{ACM MM}}.
  \bibinfo{pages}{2571--2580}.
\newblock


\bibitem[Zhu et~al\mbox{.}(2019)]%
        {zhu2019DCNv2}
\bibfield{author}{\bibinfo{person}{Xizhou Zhu}, \bibinfo{person}{Han Hu},
  \bibinfo{person}{Stephen Lin}, {and} \bibinfo{person}{Jifeng Dai}.}
  \bibinfo{year}{2019}\natexlab{}.
\newblock \showarticletitle{Deformable convnets v2: More deformable, better
  results}. In \bibinfo{booktitle}{\emph{CVPR}}. \bibinfo{pages}{9308--9316}.
\newblock


\bibitem[Zhu et~al\mbox{.}(2021)]%
        {zhu2021fce}
\bibfield{author}{\bibinfo{person}{Yiqin Zhu}, \bibinfo{person}{Jianyong Chen},
  \bibinfo{person}{Lingyu Liang}, \bibinfo{person}{Zhanghui Kuang},
  \bibinfo{person}{Lianwen Jin}, {and} \bibinfo{person}{Wayne Zhang}.}
  \bibinfo{year}{2021}\natexlab{}.
\newblock \showarticletitle{Fourier Contour Embedding for Arbitrary-Shaped Text
  Detection}. In \bibinfo{booktitle}{\emph{CVPR}}. \bibinfo{pages}{3123--3131}.
\newblock


\bibitem[Zhu et~al\mbox{.}(2016)]%
        {zhu2016fcs_survey}
\bibfield{author}{\bibinfo{person}{Yingying Zhu}, \bibinfo{person}{Cong Yao},
  {and} \bibinfo{person}{Xiang Bai}.} \bibinfo{year}{2016}\natexlab{}.
\newblock \showarticletitle{Scene text detection and recognition: Recent
  advances and future trends}.
\newblock \bibinfo{journal}{\emph{Frontiers of Computer Science}}
  \bibinfo{volume}{10} (\bibinfo{year}{2016}), \bibinfo{pages}{19--36}.
\newblock


\end{thebibliography}

%-------------------------------------------------------------------------
\newpage
\appendix

\section{More Details in Implementation}

\subsection{The Architecture of Encoder} 
\begin{figure}[!htb]
\begin{center}
\includegraphics[width=1.0\linewidth]{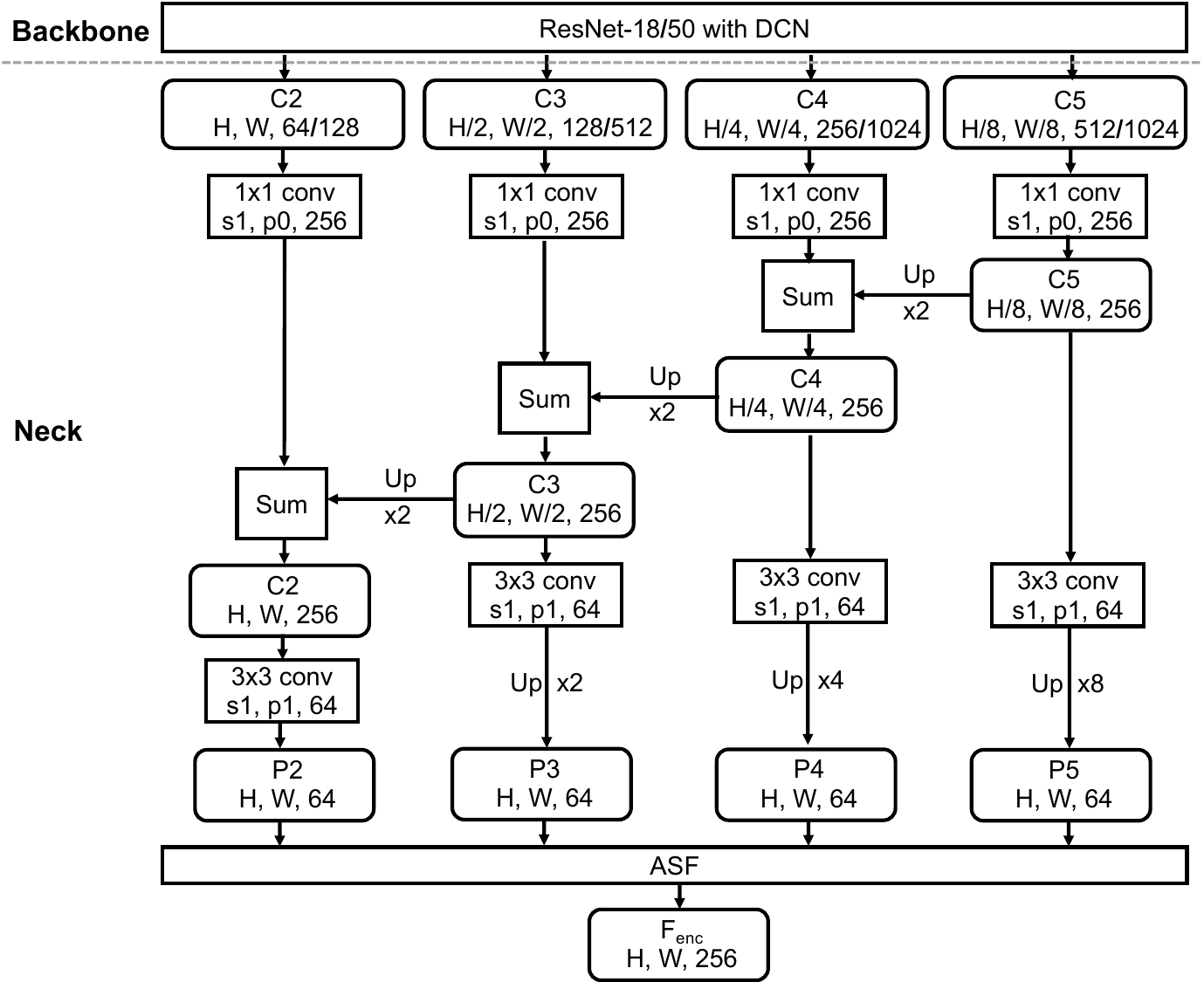}
\end{center}
\caption{The architecture of the encoder. ``s'' and ``p'' denote stride and padding of the convolution.}
\label{fig:encoder}
\end{figure}

The architecture of the encoder of SIR is exactly the same as that in \cite{liao2022db++}, which consists of the backbone part and the neck part as shown in Fig. \ref{fig:encoder}. 

\noindent\textbf{Backbone.} Following \cite{liao2022db++} and \cite{zhu2019DCNv2}, modulated deformable convolutions are applied in all the $3\times3$ convolutional layers in stages conv3, conv4, and conv5 in the ResNet backbone \cite{he2016resnet}. 

\noindent\textbf{Neck.} The neck includes a ``concat'' FPN \cite{lin2017FPN,liao2020db} and the adaptive scale fusion (ASF) \cite{liao2022db++}. Firstly, the input image is fed into an FPN to obtain features. Secondly, the pyramid features are up-sampled to the same scale and cascaded. At last, the cascaded features are fed into the ASF module \cite{liao2022db++} to produce the encoded feature $F_{enc}$.

\subsection{Anchor Setting in TDM} To handle the scale and aspect ratio variation in TDM, we construct a feature pyramid from $F_{enc}$ with four max-pooling operations, resulting in a feature pyramid of \{$F_{enc} \in {\mathbb R}^{H \times W \times C}$, $F_{enc}^{1} \in {\mathbb R}^{\frac{H}{2} \times \frac{W}{2} \times C}$, $F_{enc}^{2}\in {\mathbb R}^{\frac{H}{4} \times \frac{W }{4} \times C}$, $F_{enc}^{3} \in {\mathbb R}^{\frac{H}{8} \times \frac{W }{8} \times C} $, $F_{enc}^{4} \in {\mathbb R}^{\frac{H}{16} \times \frac{W }{16} \times C}$\}.
Vectors from different pyramids naturally encode different amounts of information on the input image, which can be associated with anchors with different scales. The aspect ratios of anchors per location are set to $\{\frac{1}{5}, \frac{1}{2}, 1, 2, 5\}$ \cite{xie2019SPCNet,liu2019BDN} since text instances are usually long.

\subsection{Label Generation}
The label generation for the probability map follows PSENet \cite{wang2019PSENet} and DB \cite{liao2020db}. Given a text image, each polygon of text instance is described by a set of segments:
\begin{equation}
G = \{S_k\}_{k=1}^{n}, 
\end{equation}
$n$ is the number of vertexes, which may vary according to different instances. Then the kernel region is generated by shrinking the polygon $G$ to $G_s$ using the Vatti clipping algorithm \cite{vatti1992vatti}. The offset $D$ of shrinking is computed from the perimeter $L$ and area $A$ of the original polygon:
\begin{equation}
D = \frac{A(1-r^2)}{L},
\end{equation}
where $r$ is the shrink ratio and is set to 0.6 empirically.
The border region is obtained by $G - G_s$, which is the area between the kernel region and the non-text region.

\subsection{Inference} 
In the inference period, the probability map is used to generate text-bounding boxes. The post-processing consists of three steps: 
(1) the probability map is firstly binarized with a constant threshold (0.65 by default) to get the binary map; 
(2) the connected regions (shrunk text regions) are obtained from the binary map by CCL; 
(3) the shrunk regions are dilated with an offset $D^{'}$ in the Vatti clipping algorithm. $D^{'}$ is calculated as:
\begin{equation}
D^{'} = \frac{A^{'} \times r^{'}}{L^{'}},
\end{equation}
where $A^{'}$ is the area of the shrunk polygon; $L^{'}$ is the perimeter of the shrunk polygon; $r^{'}$ is set to 1.3 for oriented datasets and 1.5 for curved datasets empirically. Especially, the unclip ratio is set to 0 when performing text region segmentation as the setting \# 5 in Table \ref{tab:baseline}. When evaluating the speed, we follow the way of most existing methods, in which the inference time is averaged on the whole dataset.

%-------------------------------------------------------------------------
\section{More Experimental Results}

We provide additional ablation and generalization experiments to further demonstrate the superiority of the proposed method.

\subsection{Ablation Experiments}

\noindent\textbf{The comparison of PAN \cite{wang2019PAN} loss and the proposed GDSC.} The PAN loss \cite{wang2019PAN} focuses on instance-level embedding learning which consists of pulling features from the same instance and pushing features from different instances. It is quite different from GDSC which aggregates features within the same semantic category (text) and pushes text features away from non-text that is consistent with the target of probability map learning. We supplement the result with the PAN loss on the ICDAR2015 dataset based on Table \ref{tab:ablation}, leading to Table \ref{tab:pan}. As shown in the table, when the PAN\_loss (instance embedding loss) is equipped with the segmentation baseline, a direct decrease of 3.1\% F-measure is obtained. On the contrary, GDSC brings a 1.6\% performance gain in terms of F-measure, which illustrate the superiority of the proposed GDSC. Compared with GDSC, PAN loss may fall behind for two reasons: 1) absence of consideration on text/non-text contrast; 2) the discrimination of different text instances is contradictory to text/non-text discrimination. 

\begin{table}[!htb]
\footnotesize
\centering
\caption{Detection results on ICDAR2015 with different model settings. ``P'', ``R'', and ``F'' denote precision, recall, and F-measure respectively.}
\begin{tabularx}{1.0\linewidth}{@{}lYYY@{}}
\hline
Method         & P    & R    & F    \\ \hline
Seg            & 85.0 & 77.8 & 81.2 \\
Seg + GDSC     & \textbf{87.9} & \textbf{78.2} & \textbf{82.8} \\
Seg + PAN\_loss & 85.8 & 71.6 & 78.1 \\ \hline
\end{tabularx}
\label{tab:pan}
\end{table}

\noindent\textbf{Does the mask branch \cite{he2017maskrcnn} work in TDM?} We tried the mask branch but found it does not work well. Here we provide the results on ICDAR2015, in which the settings maintain the same as those in Table \ref{tab:mask} except for model settings. As shown in Table \ref{tab:mask}, we can infer several points. Comparing the first two rows, we find that using the mask branch leads to a decrease of 3.5\% F-measure. Compared to the last two rows, adding the mask branch to TDM also brings no performance gains. We think the reason lies in two points: 1) the use of the mask branch is to provide the pixel-level representation of arbitrary-shape objects. However, via the main semantic segmentation task, the network has encoded this type of information; 2) the mask branch differentiates pixels as the foreground/background in a given bounding box. When multiple text instances lie in one box, only one instance is taken as the positive, which is opposite to the goal of semantic segmentation.

\begin{table}[!htb]
\footnotesize
\centering
\caption{Detection results on ICDAR2015 with different settings of TDM. ``Cls'' and ``Reg'' refer to the classification and regression tasks in regional recognition. ``Mask'' refers to the instance segmentation task as that in Mask R-CNN \cite{he2017maskrcnn}.}
\begin{tabularx}{1.0\linewidth}{@{}YYYYYYYY@{}}
\toprule
Seg & RPN & Cls & Reg & Mask & P    & R    & F    \\ \midrule
\checkmark   & $\times$   & $\times$   & $\times$   & $\times$    & \textbf{85.0} & 77.8 & 81.2 \\
\checkmark   & $\times$   & $\times$   & $\times$   & \checkmark    & 84.8 & 71.7 & 77.7 \\
\checkmark   & \checkmark   & \checkmark   & \checkmark   & $\times$    & 84.6 & 81.5 & \textbf{83.0} \\
\checkmark   & \checkmark   & \checkmark   & \checkmark   & \checkmark    & 83.8 & \textbf{82.1} & \textbf{83.0} \\ \bottomrule
\end{tabularx}
\label{tab:mask}
\end{table}

\subsection{Generalization Experiments}

\noindent\textbf{Generalization experiments of GDSC and TDM on other BU-SEG STD methods.} We have supplemented further experimental results with three SOTA methods upon the corresponding official implementation including DB++/DB\footnote{https://github.com/MhLiao/DB} and PSENet\footnote{https://github.com/whai362/pan\_pp.pytorch.git}, which demonstrate good generalization ability of the proposed method in this subsection.

\noindent\textbf{Experiments on DB++ \cite{liao2022db++}/DB \cite{liao2020db} .} We follow the ablation studies in DB++/DB experimenting on the MSRA-TD500 dataset and pretraining the network with SynthText. The test scale is set to 736. As shown in Table \ref{tab:db++} and Table \ref{tab:db}, the proposed GSDC and TDM consistently improve the DB++/DB baseline. Specifically, GDSC, TDM, and SIR (GDSC + TDM) respectively bring 0.6\%, 0.5\%, and 1.1\% upon DB++ and 0.6\%, 0.7\%, and 1.1\% upon DB in terms of F-measure. 

\begin{table}[!htb]
\footnotesize
\centering
\caption{Detection results on MSRA-TD500 with different model settings upon DB++.}
\begin{tabularx}{1.0\linewidth}{@{}lYYY@{}}
\toprule
Method            & P    & R    & F    \\ \midrule
DB++ (Paper)      & 87.9 & 82.5 & 85.1 \\
DB++              & \textbf{89.6} & 81.3 & 85.2 \\
DB++ + GDSC       & 88.8 & 83.0 & 85.8 \\
DB++ + TDM        & 88.2 & 83.3 & 85.7 \\
DB++ + GDSC + TDM & 88.4 & \textbf{84.4} & \textbf{86.3} \\ \bottomrule
\end{tabularx}
\label{tab:db++}
\end{table}

\begin{table}[!htb]
\footnotesize
\centering
\caption{Detection results on MSRA-TD500 with different model settings upon DB.}
\begin{tabularx}{1.0\linewidth}{@{}lYYY@{}}
\toprule
Method          & P    & R    & F    \\ \midrule
DB (Paper)      & \textbf{90.4} & 76.3 & 82.8 \\
DB              & 86.8 & 79.9 & 83.2 \\
DB + GDSC       & 88.5 & 79.6 & 83.8 \\
DB + TDM        & 88.7 & 79.6 & 83.9 \\
DB + GDSC + TDM & 86.7 & \textbf{82.0} & \textbf{84.3} \\ \bottomrule
\end{tabularx}
\label{tab:db}
\end{table}

\begin{table}[!htb]
\footnotesize
\centering
\caption{Detection results on ICDAR2015 with different model settings upon PSENet.}
\begin{tabularx}{1.0\linewidth}{@{}lYYY@{}}
\toprule
Method                         & P    & R    & F    \\ \midrule
PSENet (Official reproduction) & 83.6 & 74.0 & 78.5 \\
PSENet                         & 83.2 & 74.3 & 78.5 \\
PSENe + GDSC                   & 82.8 & 74.0 & 78.2 \\
PSENe + TDM                    & 86.4 & \textbf{79.5} & 82.8 \\
PSENe + GDSC + TDM             & \textbf{88.9} & 79.1 & \textbf{83.7} \\ \bottomrule
\end{tabularx}
\label{tab:psenet}
\end{table}

\noindent\textbf{Experiments on PSENet \cite{wang2019PSENet}}. The experiments are performed on the ICDAR2015 dataset without pretraining. The short side is set to 736 pixels during evaluation. As shown in Table \ref{tab:psenet}, several interesting points can be concluded: (1) Only using GDSC doesn't help PSENet; (2) TDM brings a significant increment of 4.3\% F-measure upon the baseline; (3) Adding GDSC and TDM together brings additional 0.9\% than TDM only. We find the phenomenons originate from the specific designs of PSENet which learn muti-scale text kernels and use progressive scale expansion for post-processing. The reason for point (1) is that learning multi-scale kernels may introduce fussy semantics of foreground/background for GDSC. For point (2), besides the errors from learned multi-scale kernels, the progressive scale expansion further amplifies the effect due to the accumulated errors, making PSENet benefit greatly from the top-down instance-level information brought by TDM. Point (3) demonstrates the complementarity of GDSC and TDM; GDSC performs as a global semantic regularizer for TDM. 

%-------------------------------------------------------------------------
\section{More Visualization Results}
In this section, for better illustration, we provide more visualization results including the BU-SEG baseline (Fig. \ref{fig:base_ic15} and \ref{fig:base_tt}), and comparison of the baseline and SIR (Fig. \ref{fig:more_ab}), and more detection results (Fig. \ref{fig:vis_det}). 

\subsection{The BU-SEG Baseline}
\begin{figure*}[!htb]
\begin{center}
\includegraphics[width=1.0\linewidth]{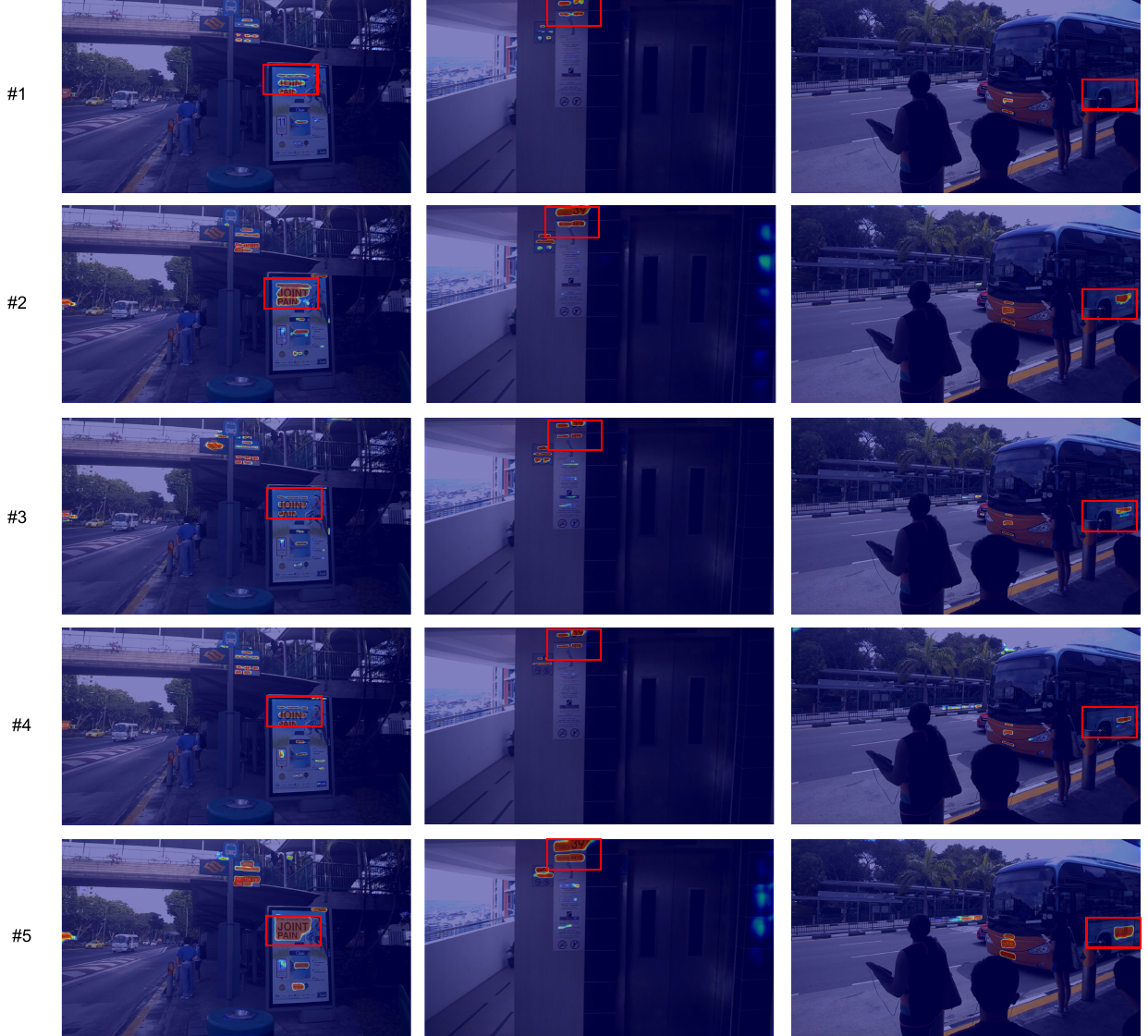}
\end{center}
\caption{Heatmaps generated with the probability maps and the original images on the ICDAR2015 dataset. \#1-\#5 rows are the corresponding results produced by the settings \#1-\#5 in Table 1 of the main body. \#1: ``kernel''; \#2: ``kernel+W\_kn''; \#3: ``kernel+W\_kbn''; \#4: ``kernel+W\_kbn+border''; \#5: ``text+W\_kbn''.}
\label{fig:base_ic15}
\end{figure*}

\begin{figure*}[!htb]
\begin{center}
\includegraphics[width=1.0\linewidth]{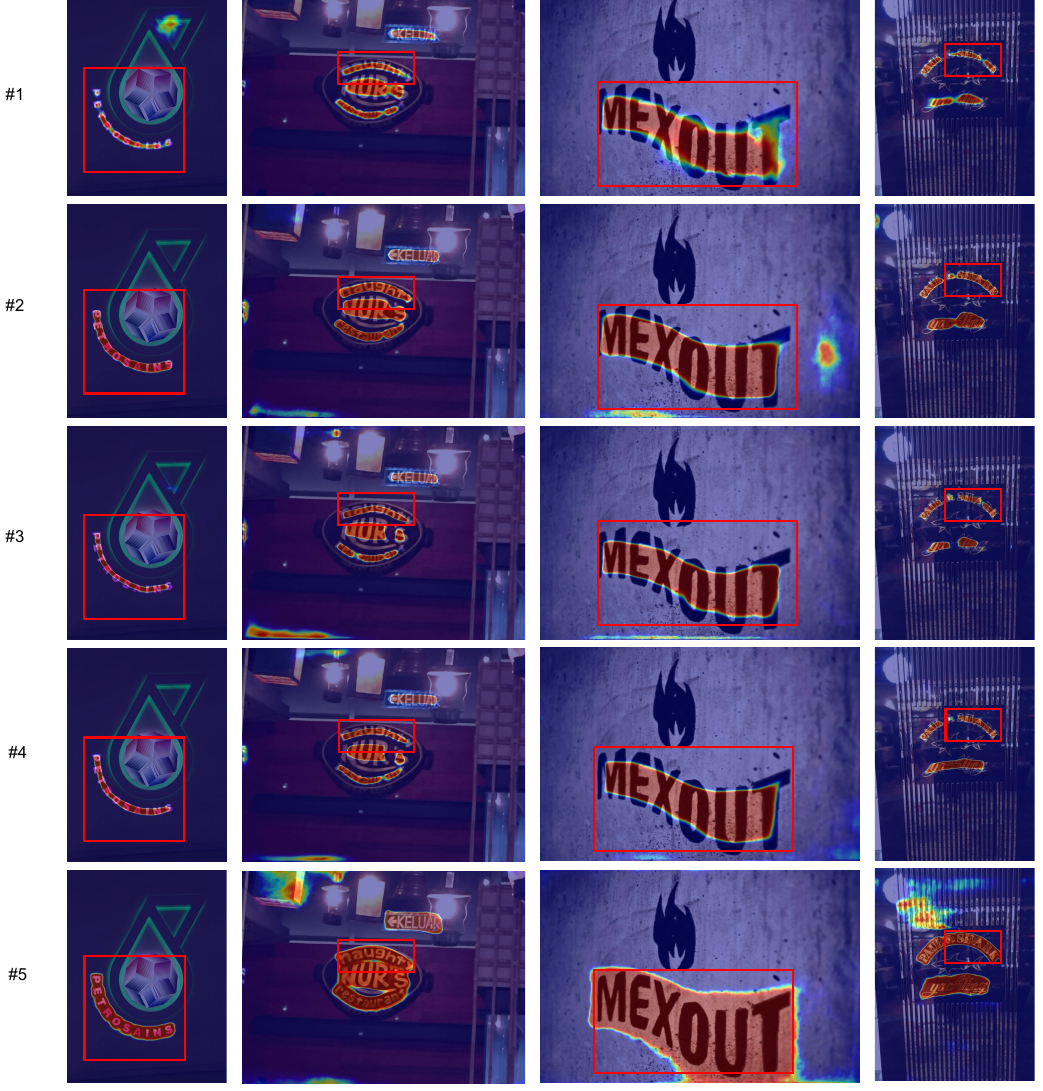}
\end{center}
\caption{Heatmaps generated with the probability maps and the original images on the Total-Text dataset. \#1-\#5 rows are the corresponding results produced by the settings \#1-\#5 in Table 1 of the main body. \#1: ``kernel''; \#2: ``kernel+W\_kn''; \#3: ``kernel+W\_kbn''; \#4: ``kernel+W\_kbn+border''; \#5: ``text+W\_kbn''.}
\label{fig:base_tt}
\end{figure*}

As shown in Fig. \ref{fig:base_ic15} and \ref{fig:base_tt}, among settings from \#1 to \#5 in Table \ref{tab:baseline}, \#4 which performs a three-category classification weighted by class-aware coefficients performs best. 
We can find that the kernel boundary is not entirely clear in \#2 due to an imbalance of samples around the kernel boundary.
When comparing \#3 to \#2, the weights of samples in the border region are raised to the same level as the kernel region, which leads to a more balanced learning process and better kernel boundary. 
When comparing \#4 to \#3, the border and non-text classes are decoupled to be supervised independently, which introduces explicit semantics and makes the kernel boundary more smooth. Looking at the results of \#5, which performs text segmentation, we find the learned probability maps are also reasonable, demonstrating the generalization of the proposed BU-SEG segmentation baseline.

\subsection{Comparison of the Baseline and SIR}
\begin{figure*}[!htb]
\begin{center}
\includegraphics[width=1.0\linewidth]{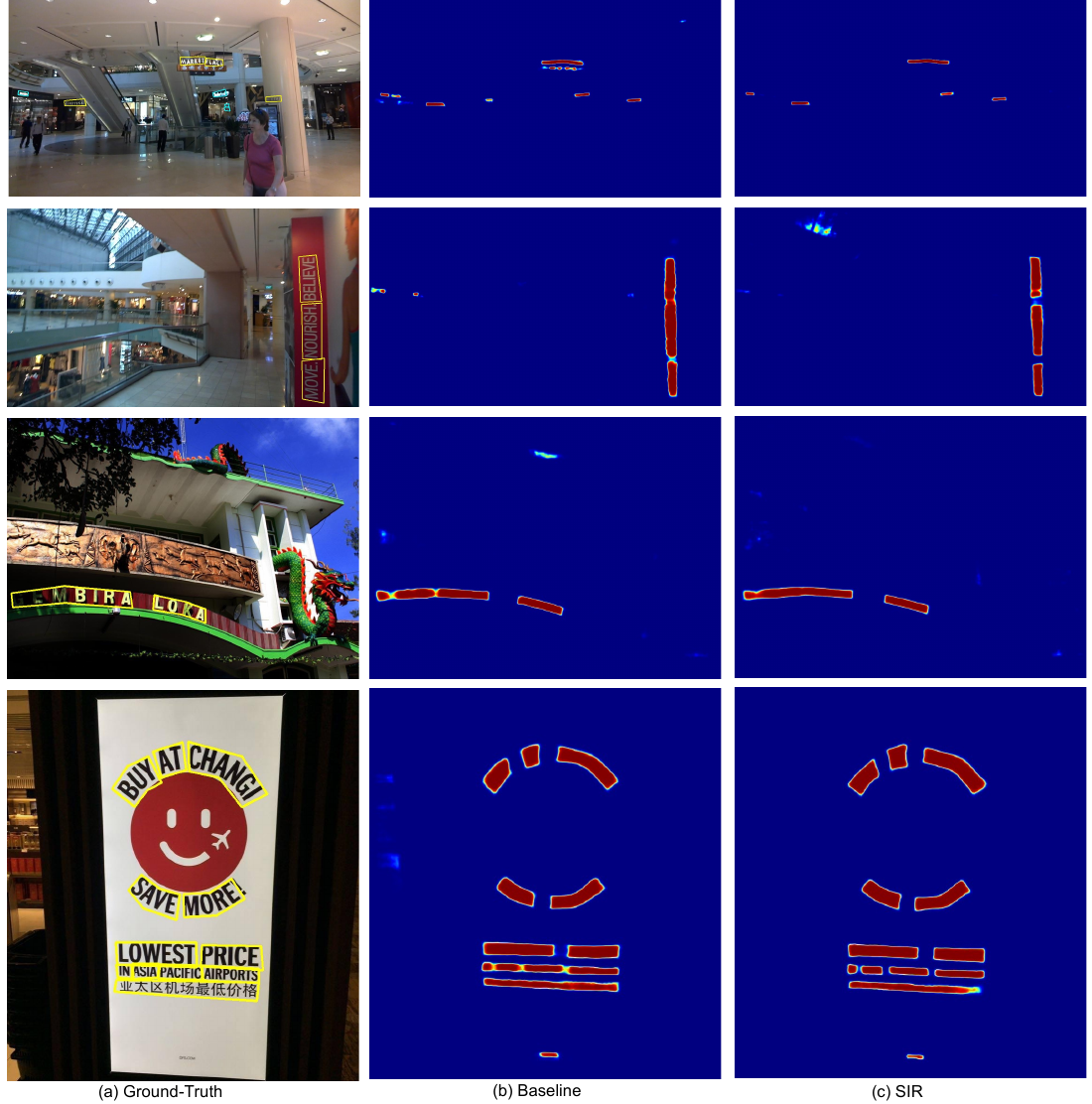}
\end{center}
\caption{Visualization results of the probability maps on ICDAR2015 and Total-Text. (b) and (c) denote the probability maps generated by the baseline segmentation model and SIR respectively; (a) denotes the ground truth. SIR improves the quality of the probability map, which can produce more robust detection results.}
\label{fig:more_ab}
\end{figure*}

In this section, we provide more samples of the probability maps produced by the baseline model and SIR on the ICDAR2015 and Total-Text datasets. As shown in Fig. \ref{fig:more_ab}, with the proposed auxiliary modules and tasks, the prediction results are more robust against false positives and text adhesion. The network learns more discriminative features on text/non-text classification and nearby text discrimination. SIR improves the quality of the probability maps, which can produce more robust detection results.

\subsection{Detection Results}

\begin{figure*}[!htb]
\begin{center}
\includegraphics[width=1.0\linewidth]{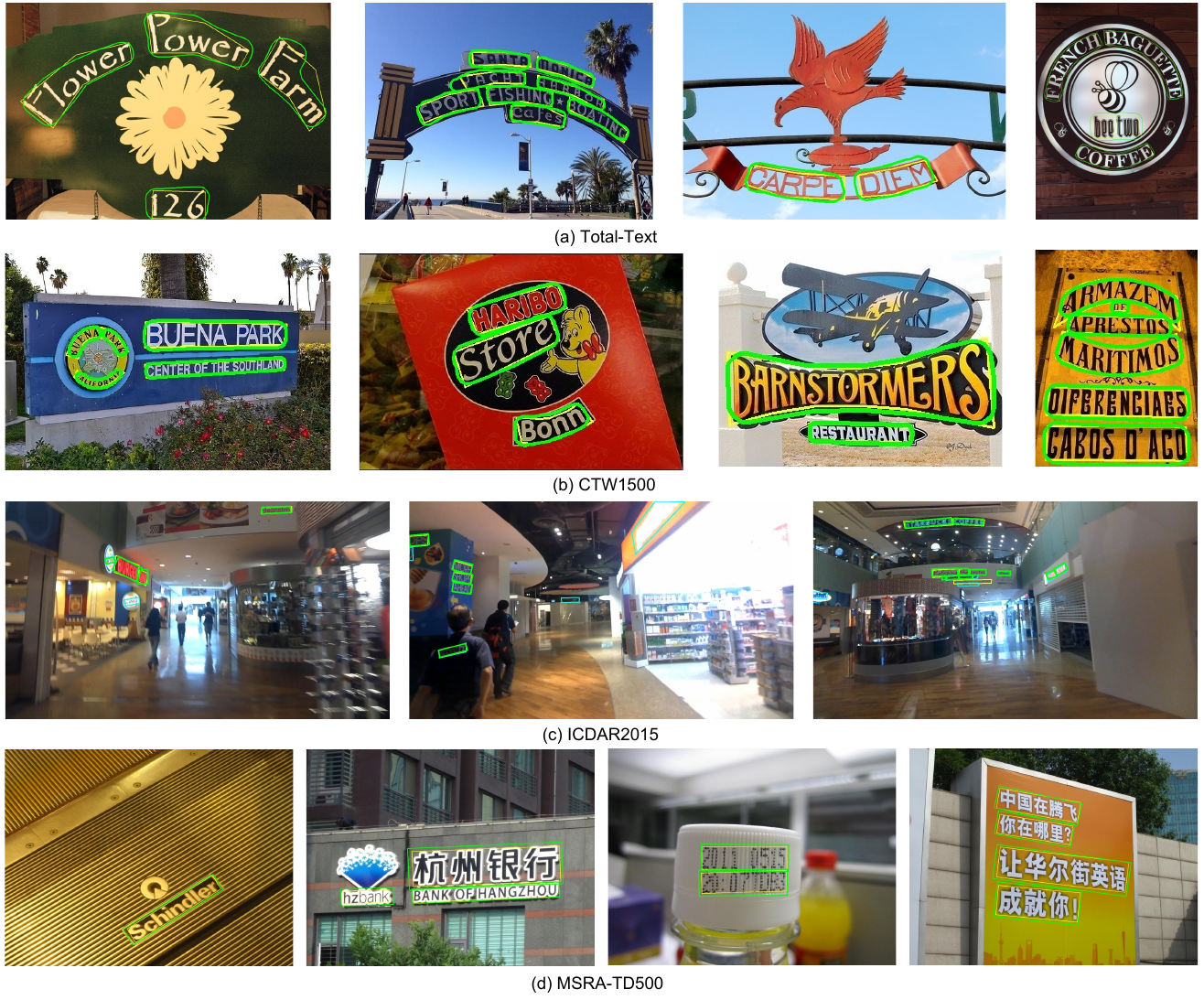}
\end{center}
\caption{Qualitative results of SIR in different cases on the Total-Text, CTW1500, ICDAR2015, and MSRA-TD500 datasets, including multi-oriented text, long text, multi-lingual text, low-resolution text, curved text, and dense text. The green and yellow boxes denote the detection results and the ground truth respectively. The cyan color represent ``do-not-care'' region.}
\label{fig:vis_det}
\end{figure*}

As shown in Fig. \ref{fig:vis_det}, we provide more qualitative results of SIR in different cases on the Total-Text, CTW1500, ICDAR2015, and MSRA-TD500 datasets, including multi-oriented text, long text, multi-lingual text, low-resolution text, curved text, and dense text, which illustrate the robustness of SIR on detecting various kinds of scene text.

%-------------------------------------------------------------------------

\end{document}